\documentclass{ieeetj}
\usepackage{cite}
\usepackage{amsmath,amssymb,amsfonts}
\usepackage{graphicx,color}
\usepackage{textcomp}
\usepackage{xcolor}
\usepackage{hyperref}
\hypersetup{hidelinks=true}

\def\BibTeX{{\rm B\kern-.05em{\sc i\kern-.025em b}\kern-.08em
    T\kern-.1667em\lower.7ex\hbox{E}\kern-.125emX}}
\AtBeginDocument{\definecolor{tmlcncolor}{cmyk}{0.93,0.59,0.15,0.02}\definecolor{NavyBlue}{RGB}{0,86,125}}

\usepackage{setspace}

\usepackage{afterpage}

\usepackage[utf8]{inputenc}
\usepackage{glossaries}
\usepackage{xspace} 

\usepackage{amsmath,amsfonts}
\usepackage{array}
\usepackage [english]{babel}
\usepackage [autostyle, english = american]{csquotes}
\MakeOuterQuote{"}

\usepackage{textcomp}
\usepackage{stfloats}
\usepackage{url}
\usepackage{verbatim}
\usepackage{graphicx}

\usepackage[numbers]{natbib} 
\usepackage{balance}

\usepackage{layouts} 
\usepackage{times}
\usepackage{algorithm2e}

\usepackage[section]{placeins}
\usepackage{leftindex}
\usepackage{todonotes}
\hypersetup{bookmarks=true,colorlinks=true,citecolor=blue,linkcolor=blue}
\usepackage{booktabs} 

\usepackage{mathtools}
\usepackage{multicol}
\usepackage{annotate-equations} 
\usepackage{xcolor}
\usepackage{lipsum} 
\usepackage{afterpage}
\usepackage{cuted}
\usepackage{float}
\usepackage{siunitx}
\usepackage{xspace}
\usepackage{threeparttable}
\usepackage{tablefootnote}
\usepackage[abbreviations]{glossaries-extra}

\definecolor{magenta_dark}{RGB}{149,35,144}
\definecolor{commentcolor}{RGB}{0,128,0}
\definecolor{keywordcolor}{RGB}{0,0,255}
\definecolor{functioncolor}{RGB}{128,0,128}

\SetCommentSty{mycommfont}

\usepackage{caption}
\usepackage{subcaption}
\usepackage{etoolbox} 

\newcommand{\ra}[1]{\renewcommand{\arraystretch}{#1}}

\def\secref#1{Sec.~\ref{#1}}

\def\figref#1{Fig.~\ref{#1}}
\def\tabref#1{Tab.~\ref{#1}}
\def\eqref#1{Eq.~(\ref{#1})}

\glssetcategoryattribute{acronym}{indexonlyfirst}{true}

\newcommand{\pose}[3]{\mathbf{T}_{\mathtt{#1 #2}}^{#3}}

\newcommand{\rot}[3]{\mathbf{R}_{\mathtt{#1 #2}}^{#3}}
\newcommand{\pos}[2]{{\mathtt{_#1}} \mathbf{p}^{#2}}

\newcommand{\frameW}{\mathtt{W}} 
\newcommand{\frameL}[1]{\leftindex{^{#1}}\mathtt{L}} 
\newcommand{\frameC}[1]{\leftindex{^{#1}}\mathtt{C}} 
\newcommand{\frameB}{\mathtt{B}} 
\newcommand{\frameBG}{\mathtt{B}_{g}} 

\newcommand{\K}[1]{\leftindex{^#1}\mathbf{K}_{3\times3}}
\newcommand{\img}[3]{\leftindex{^#1}\mathbf{I}_{#2}^{#3}}

\newcommand{\pcd}[3]{\leftindex{^#1}\mathbf{P}_{#2}^{#3}}
\newcommand{\pcdm}[2]{\mathbf{P}_{#1}^{#2}}

\newcommand{\feat}[4]{\mathbf{F}_{#1 \times #2 \times #3}^{#4}}

\newcommand{\grid}[2]{\mathbf{G}_{#1}^{#2}}
\newcommand{\gridhat}[2]{\mathbf{\hat{G}}_{#1}^{#2}}

\newcommand{\loss}[1]{\mathcal{L}_{\mathrm{#1}}}

\newcommand{\fun}[2]{f_{\mathrm{#1}}\left( #2 \right) }

\robustify{\pose}
\robustify{\rot}
\robustify{\pos}
\robustify{\K}
\robustify{\loss}
\robustify{\feat}
\robustify{\img}
\robustify{\fun}

\usepackage{xspace}
\makeatletter
\DeclareRobustCommand\onedot{\futurelet\@let@token\@onedot}
\def\@onedot{\ifx\@let@token.\else.\null\fi\xspace}
\makeatother

\usepackage[abbreviations]{glossaries-extra}

\glssetcategoryattribute{abbreviation}{indexonlyfirst}{true}

\glssetcategoryattribute{abbreviation}{nohyperfirst}{true}

\newcommand{\grandchallenge}{DARPA Grand Challenge\xspace}

\newabbreviation{auroc}{AUROC}{Area Under the Receiver Operating Characteristic Curve}
\newabbreviation{accuracy}{Acc}{Accuracy}

\newabbreviation{bev}{BEV}{Bird`s Eye View}
\newabbreviation{camera}{}{Carnegie Robotics MultiSense S27}
\newabbreviation{cnn}{CNN}{Convolutional Neural Network}
\newabbreviation{cvar}{CVaR}{Conditional Value at Risk}

\newabbreviation{dem}{DEM}{Digital Elevation Model}

\newabbreviation{fov}{FoV}{Field of View}

\newabbreviation{gnn}{GNN}{Graph Neural Network}
\newabbreviation{gcn}{GCN}{Graph Convolutional Network}
\newabbreviation{hdd}{HDD}{Hazard Distance Detection}
\newabbreviation{imu}{IMU}{Inertial Measurement Unit}
\newabbreviation{irl}{IRL}{Inverse Reinforcement Learning}

\newabbreviation{knn}{KNN}{K-Nearest Neighbors}

\newabbreviation{lagr}{LAGR}{Learning Applied to Ground Vehicles}
\newabbreviation{lidar}{LiDAR}{Light Detection and Ranging}
\newabbreviation{ladar}{LADAR}{LAser Detection And Ranging}
\newabbreviation{lidarsensor}{}{Velodyne VLP-32C}
\newcommand{\lidar}{LiDAR\xspace}

\newabbreviation{mlp}{MLP}{Multi-Layer Perceptron}
\newabbreviation{mpc}{MPC}{Model Predictive Controller}
\newabbreviation{mse}{MSE}{Mean Squared Error}
\newabbreviation{mae}{MAE}{Mean Absolute Error}
\newabbreviation{mppi}{MPPI}{Model Predictive Path Integral}

\newabbreviation{nir}{NIR}{Near-InfraRed}

\newabbreviation{wmse}{WMSE}{Weighted Mean Squared Error}
\newabbreviation{wmae}{WMAE}{Weighted Mean Absolute Error}

\newabbreviation{ours}{RoadRunner}{RoadRunner}

\newabbreviation{rbf}{RBF}{Radial Basis Function}
\newabbreviation{rmp}{RMP}{Riemannian Motion Policies}
\newabbreviation{ros}{ROS}{Robot Operating System}
\newabbreviation{ros1}{ROS~1}{Robot Operating System}
\newabbreviation{roc}{ROC}{Receiver Operating Characteristic}
\newabbreviation{rf}{RF}{Random Forest}
\newabbreviation{radar}{RADAR}{RAdio Detection And Ranging}
\newabbreviation{rl}{RL}{
Reinforcement Learning}

\newabbreviation{sdf}{SDF}{Signed Distance Field}
\newabbreviation{slam}{SLAM}{Simultaneous Localization and Mapping}
\newabbreviation{svm}{SVM}{Support Vector Machine}
\newabbreviation{svc}{SVC}{Support Vector Classifier}
\newabbreviation{stack}{X-Racer}{X-Racer}

\newabbreviation{work}{TraFo}{TerrainFormer}

\newabbreviation{ugv}{UGV}{Unmanned Ground Vehicle}

\newabbreviation{wvn}{WVN}{Wild Visual Navigation}

\newabbreviation{vit}{ViT}{Vision Transformer}
\newabbreviation{vehicle}{}{Polaris RZR S4 1000 Turbo}

\def\OJlogo{\vspace{-4pt}}
\def\seclogo{\vspace{10pt}}

\def\authorrefmark#1{\ensuremath{^{\textbf{#1}}}}

\begin{document}
\receiveddate{XX Month, XXXX}
\reviseddate{XX Month, XXXX}
\accepteddate{XX Month, XXXX}
\publisheddate{XX Month, XXXX}
\currentdate{XX Month, XXXX}
\doiinfo{XXXX.2022.1234567}

\markboth{}{Frey {et al.}}

\title{RoadRunner - Learning Traversability Estimation for Autonomous Off-road Driving}

\author{Jonas Frey\authorrefmark{1,}\authorrefmark{2}, Manthan Patel\authorrefmark{1,}\authorrefmark{2}, Deegan Atha \authorrefmark{1}, Julian Nubert \authorrefmark{1,}\authorrefmark{2}, David Fan\authorrefmark{1}, Ali Agha\authorrefmark{1}, Curtis Padgett\authorrefmark{1}, Patrick Spieler\authorrefmark{1}, Marco Hutter\authorrefmark{2}, Shehryar Khattak\authorrefmark{1}}
\affil{Jet Propulsion Laboratory (JPL), California Institute of Technology (Caltech), Pasadena, CA 91011, USA}
\affil{Swiss Federal Institute of Technology (ETH Z\"urich), Robotic Systems Lab, Z\"urich 8092, Switzerland}
\corresp{Corresponding author: Jonas Frey (email: jonfrey@ ethz.ch).}
\authornote{The research was carried out at the Jet Propulsion Laboratory, California Institute of Technology, under a contract with the National Aeronautics and Space Administration (80NM0018D0004). This work was partially supported by the Defense Advanced Research Projects Agency (DARPA) and the Swiss Federal Institute of Technology, ETH Zurich. Jonas Frey and Julian Nubert are supported by the Max Planck ETH Center for Learning Systems.}

\begin{abstract}
Autonomous navigation at high speeds in off-road environments necessitates robots to comprehensively understand their surroundings using onboard sensing only. The extreme conditions posed by the off-road setting can cause degraded image quality as well as limited sparse geometric information available from LiDAR sensing when driving at high speeds. In this work, we present RoadRunner, a novel framework capable of predicting terrain traversability and elevation directly from camera and LiDAR sensor inputs. RoadRunner enables reliable autonomous navigation by fusing sensory information and generates contextually informed predictions about the geometry and traversability of the terrain while operating at low latency. In contrast to existing methods, which rely on classifying handcrafted semantic classes and using heuristics to predict traversability costs, our method directly predicts traversability. It is trained on labels that can be automatically generated in hindsight in a self-supervised fashion. The RoadRunner network architecture builds upon advances from the autonomous driving domain, which allow us to embed LiDAR and camera information into a common Bird`s Eye View perspective. Training is enabled by utilizing an existing traversability estimation stack to generate training data in hindsight in a scalable manner from real-world off-road driving datasets. Furthermore, RoadRunner improves the system latency by a factor of $\sim4$, from $\SI{500}{ms}$ to $\SI{140}{ms}$, while improving the accuracy for traversability costs and elevation map predictions. We demonstrate the effectiveness of RoadRunner in enabling safe and reliable off-road navigation at high speeds in multiple real-world driving scenarios through unstructured desert environments.
\end{abstract}


\maketitle

\section{INTRODUCTION}
\label{sec:Introduction}
\begin{figure*}[t]
    \centering
    \includegraphics[width=1.0\textwidth]{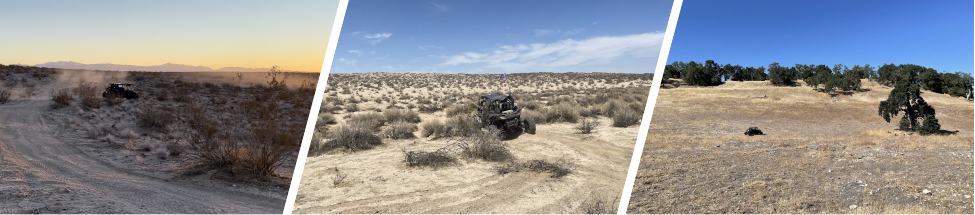}
    \caption{Example deployment environments showcasing high-speed off-road navigation. }
    \label{fig:introduction-offroad_navigation}
\end{figure*}
\IEEEPARstart{A}{nimals}, like the greater roadrunner, are capable of traversing complex off-road terrains at an impressive running speed of \SI{24}{km/h} to \SI{32}{km/h} \cite{maxon2005real}.
This capability makes the roadrunner a role model in terms of perception capabilities required for autonomous ground vehicles to advance toward enabling high-speed operations in unstructured off-road environments.
One of the key components to facilitate safety while driving at high speeds is the assessment of where a robot can navigate safely. This assessment is known as  \emph{traversability estimation}, where the objective is to estimate the affordance or risk the robot must take when navigating the perceived terrain.
The affordance may depend on the robot's hardware, control system, terrain geometry, physical terrain properties, and other factors.

Complex and unstructured off-road environments pose a variety of obstacles and other potential risks, which need to be correctly assessed by an autonomous ground vehicle. 
In addition, short reaction times are required at high speeds and potential hazards must be identified at long distances to guarantee safety. 
Moreover, only partial observations of the environment are available, given that onboard sensors like LiDARs and cameras can only provide information at a limited update rate, and only perceive a restricted field of view with increasing intrinsic sparsity at longer distances. 
Additionally, when considering driving off-road, robots cannot rely on high-definition map information, given that the robot may be deployed for the first time in the environment (space exploration) or the environment might have changed significantly due to seasonal variations.
Furthermore, extreme conditions such as dust, dirt, fog, and rain are common in an off-road setting and can lead to degraded perception modalities, as shown in the case of \lidar sensors~\cite{stanislas2021airborne}.
As a result, the traversability estimation system has to cope with all of these challenges to allow for navigation in an off-road setting at high speeds.

In comparison, self-driving cars operate at higher maximum speeds and in more dynamic environments. Despite this, the terrain traversability estimation remains simple as cars primarily drive on paved roads.
Therefore, the classification of scene semantics is mostly sufficient to conclude the traversability.
As a result, most works focus on identifying semantic classes like lanes, cars, and traffic signs from images~\cite{Philion2022} or \lidar data~\cite{Lang2019}. Despite the difference in the use case, self-driving cars are equipped with a similar sensing setting, and our work is inspired by the recent advances in multi-modal network architectures that fuse different sensory information into a common \gls{bev} perspective~\cite{Philion2022,Liu2022}.

Historically, the first progress towards autonomous navigation in unstructured off-road environments was achieved throughout the DEMO III~\cite{Shoemaker1998}, PerceptOR~\cite{Kelly2006}, and \gls{lagr}~\cite{Kim2006,Hadsell2009} programs, as well as the \grandchallenge. As part of these programs, multiple works pioneered traversability estimation from camera and \lidar data and applied learning-based approaches, as well as, non-learning-based approaches, which rely on heuristics to assess the terrain traversability. 
More recently, using simulation-based approaches~\cite{Guzzi20,Frey2022}, which can evaluate the terrain with respect to the specific robot's capabilities, and self-supervised approaches, which make use of real-world deployment data~\cite{Frey2023} have gained significant popularity, for estimating traversability in an off-road setting.

To further enhance the off-road traversability estimations, we present \glsentrylong{ours}: a multi-camera multi-\lidar learning architecture able to predict terrain traversability and elevation with low latency directly from sensor inputs. 
Our architecture leverages point cloud and pre-trained image segmentation backbones and fuses information into a unified \gls{bev} representation.
A \gls{cnn} predicts a robot-centric traversability and elevation map from the unified feature embedding. 
The training data is generated in a self-supervised manner using hindsight to determine the traversability and the geometry of the terrain without any human labeling or explicit identification of semantic classes during run-time. 
The concept of hindsight refers to the fact that offline processing with future and past sensor measurements can be used to create more reliable estimates compared to the online deployment of the same algorithm.
To generate the training data for \glsentrylong{ours} we use NASA Jet Propulsion Laboratory offroad autonomy research stack (\glsentrylong{stack}) (\secref{sec:method-preliminary}), a sophisticated off-road driving stack that includes mapping and traversability estimation, which leverages geometry and semantics with carefully crafted and field-proven heuristics.  These heuristics encode the vehicle-specific capabilities. While \glsentrylong{stack} is not the main contribution of this paper, we firstly provide an overview of all relevant components to this work within \secref{sec:method-preliminary}. 
\glsentrylong{ours} provides a framework for using hindsight and end-to-end learning to overcome the limitations inherent to a non-learned traversability software stack. 
In addition, by predicting traversability and elevation simultaneously, \glsentrylong{ours} functions as a single perception and mapping module, providing all perceptual information necessary for downstream path planning. Throughout this work, all experiments are conducted using our customized \glsentrylong{vehicle}. This side-by-side off-road vehicle functions as a testbed to enhance off-road autonomy and can be seen in \figref{fig:introduction-offroad_navigation}. 

The key contributions of our work are:
\begin{itemize}[noitemsep]
    \item A novel \glsentrylong{ours} network architecture that can simultaneously predict traversability costs and elevation map information directly from multi-LiDAR and multi-camera data at low latency.
    \item A general framework to generate pseudo-ground truth elevation maps and traversability costs using temporal data aggregation in hindsight to train the network in a self-supervised manner. 
    \item We present an overview of NASA Jet Propulsion Laboratory's off-road autonomy research stack \glsentrylong{stack}, one of the most advanced off-road autonomy software stacks.
    \item An exhaustive evaluation of the proposed  \glsentrylong{ours} architecture on real-world field test data and comparisons to existing network architectures as well as an ablation study.
\end{itemize}

\glsentrylong{ours} outperforms the elevation mapping and traversability estimation of \glsentrylong{stack}, which was used to generate training data, by leveraging visual and geometric data. In addition, it is capable of predicting missing elevation and traversability information based on the learned context.
The same holds for traversability detection, where obstacles can be detected at a longer range.
We demonstrated that our proposed RoadRunner architecture, compared to the X-Racer stack, improves traversability cost estimation by \SI{52.3}{\%} in MSE and \SI{36.0}{\%} in MAE for elevation map estimation while reducing the perception-to-traversability latency during inference by a factor of $\sim4$ compared to the existing software stack.

\section{RELATED WORK}
\label{sec:Related work}
In this section, we provide an overview of traversability estimation methods, focusing on off-road driving, traversability from semantics, self-supervision, and other approaches. In addition, we provide an overview of semantic segmentation methods operating in a BEV representation. 

\subsection{Traversability Estimation for Off-Road Driving}

Terrain traversability is dependent upon both the terrain's geometry and its physical properties like stiffness, friction, or granularity. 
Several other factors influence the traversability, which include the specific robotic system in use, the applied control strategy~\cite{Frey2023}, and the robot's state during its interaction with the terrain~\cite{Fan2021}.
For example, different robots (\emph{hardware and control dependence}) can surmount distinct obstacles when driving at varying speeds (\emph{state dependence}).
In this work, we simplify this multi-variant view of traversability following~\cite{Fan2021} and model the traversability as the affordance/risk required to overcome a given terrain as a probability distribution. 
We marginalize the robot-state dependency and target a specific robotic system and control strategy. 
To make the overall traversability more interpretable, we employ the \gls{cvar} metric.
The \gls{cvar} metric, which is a \textit{coherent risk metric} being monotonic, subadditive, homogeneous, and translational invariant, allows us to emphasize different parts of the risk distribution associated with a given terrain~\cite{Majumdar2017}.

\cite{Pomerleau1993} pioneered learning for autonomous driving by training a \gls{mlp} to predict online the steering command from a low-resolution camera image by learning from demonstration for on-road driving end-to-end. 
The first steps towards off-road autonomy were achieved in the DEMO III project~\cite{Shoemaker1998} followed by the Perceptor program~\cite{Kelly2006}, in which off-road capable \glspl{ugv} were retrofitted with a variety of sensors, including cameras and LiDARs for cross-country navigation.
The developed perception system at the time was primarily based on heuristics and, therefore failed to generalize to the complex off-road domain. 
As part of this program~\cite{Manduchi2005} classified terrain into a discrete set of classes from color images and used stereo depth to detect obstacles, highlighting the importance of using different sensor modalities. 

The \gls{lagr} program~\cite{Jackel2006} aimed to improve the perception system developed within the Perceptor program using machine learning techniques. 
For this,~\cite{Muller2005} expanded on the idea of predicting the steering command end-to-end from images~\cite{Pomerleau1993}, but instead of an \gls{mlp} used a \gls{cnn} in an off-road setting. For the network to learn correct driving behavior, it implicitly learned obstacle avoidance, and thereby implicitly traversability. 
\cite{Hadsell2007} used reliable short-range geometric measurements to supervise the training of a simple classifier to predict the traversability from image data in a self-supervised manner and proposed the concept of spatial label propagation. 
\cite{Kim2006} instead clustered the environment and assigned positive and negative labels through interaction to determine the traversable areas. 
\cite{Konolige2009} explicitly identified the ground plane from stereo depth and classified paths based on image segmentation and geometric information by fitting a model from a single image before each deployment.

As part of the \grandchallenge \cite{Behringer2004}, ~\cite{Thrun2006} proposed a probabilistic terrain analysis for high-speed desert driving taking into account \lidar measurements which are used to assess the obstacles based on height difference in combination with an adaptive vision approach.
In contrast,~\cite{Urmson2006} fully relied on geometry captured by \lidar and RADAR. Both teams showcased long-term autonomy in a desert scenario, within a narrow scope and perfectly pre-planned routes under GPS guidance. Our overall system does not require precise GPS guidance and can drive in more complex open-field environments at high speeds. 

\subsubsection{Traversability from Semantics} 
Modern deep learning methods excel in semantic scene understanding from image-~\cite{Tan19,Viswanath2021,Maturana2017,Roth2023} and point cloud data~\cite{Shaban2022}. 
Commonly, identified semantic classes are fused into a map representation~\cite{Gian2023,Maturana2017} and then associated with a traversability score~\cite{Maturana2017} using heuristics about the traversability properties of the given semantic class. 
Notably, datasets like RUGD~\cite{Wigness2019rugd}, Rellis-3D~\cite{Jiang2021rellis} or Freiburg Forest~\cite{Valada16freiburg} provide high-quality annotated off-road semantics, however, are limited in size and diversity, hindering generalization to diverse off-road scenarios.
Moreover, the reliance on manual labeling and heuristic-based mapping from semantics to traversability inherently limits the performance of the traversability estimation.

Within an off-road setting,~\cite{Bradley2015} perform a per-voxel terrain density classification from \lidar and camera images using a \gls{rf} classifier on a diverse legged robot dataset. 
\cite{Maturana2017} perform semantic segmentation of images and project the labels into a 2.5D semantic elevation map for a \gls{ugv}. 
\cite{Schilling2017} utilize learned semantics segmentation features and heuristic features from \lidar data to predict safe, risky, and obstacle regions.
\cite{Viswanath2021} predict semantics in image space using a \gls{cnn} while~\cite{Shaban2022} predict semantics in the \gls{bev} perspective using \lidar. \cite{Shaban2022} employ a sparse \gls{cnn} and allow to feed-forward latent features from previous predictions in combination with memory units.

Rovers used in space exploration face the same perception challenges as off-road driving, where the accurate identification of terrain hazards is pivotal to mission success~\cite{Ono2015}. As a result, the identification of semantic terrain classes and physical interaction has been exhaustively studied.
\cite{Rothrock2016} employed a \gls{cnn} to classify Martian terrain and predict wheel slip of the Mars Perseverance rover. 
\cite{Swan2021} curated a comprehensive Mars terrain semantic segmentation dataset. Other works focus on improving semantic segmentation by reducing the number of training samples, given the challenges associated with collecting training data in a space exploration scenario~\cite{Zhang2022,Goh2022}.
\cite{Endo2023} assess risk-aware traversability costs by fusing semantic terrain classification and a slip model in a probabilistic manner.

Most relevant to our work is \textit{TerrainNet}~\cite{Meng2023}, which follows~\cite{Triebel2006} by modeling the environment by a ground and ceiling layer, where each layer contains its associated semantics. 
The geometry and semantics are predicted by a neural network taking multiple RGB and depth camera images as input.
Features from the camera are accumulated in a \gls{bev} representation by predicting the corresponding location of the feature in image space from stereo depth.

\subsubsection{Traversability from Self-Supervision} 
Traversability estimation approaches based on scene semantics typically require expensive annotated data. 
Methods operating in a self-supervised manner aim to overcome this limitation by generating a training signal without relying on manual annotation. 
Instead they exploit information from other sensor modalities~\cite{Brooks2012,Otsu2016,Castro2023,Seo2023learning,Higa2019,Meng2023,Zurn2021,Sathyamoorthy2022}, or the interaction of the robot with the environment~\cite{Richter17,Seo2022,Frey2023,Ahtiainen2017,Gasparino2022,Cai2022,Xue2023contrastive,Sathyamoorthy2022,Cai2023evora,Jung2023}. 
The generated supervision signal allows training a model that predicts a look-ahead estimate of the terrain, all without requiring the robot to be near to or interact with the terrain.

\cite{Brooks2012} train an image classifier to predict terrain classes of the Mars-analogue environment identified by the proprioception.
\cite{Otsu2016} use co-and self-training to improve two classifiers to predict terrain classes from images and proprioception for space exploration. 
\cite{Ahtiainen2017} learn traversability from \lidar measurements by accumulating information in a map and training a \gls{svm} to classify traversability supervised by positive labels the robot visited and negative labels based on heuristics.
\cite{Higa2019} directly estimate driving energy for rovers from images using the measured energy consumption. 
\cite{Castro2023} adapt the proprioception-based pseudo labels by measuring vibration data, which is used as the network input to predict a traversability grid map from a colorized elevation map. 

\cite{Richter17} proposed to use anomaly detection to predict safe image regions for indoor navigation with a wheeled robot.
Multiple other works integrated the concept of anomaly detection and tools available in evidential deep learning to learn from real-world data without manual labeling~\cite{Frey2023,Schmid2022,Seo2022,Wellhausen2020,Cai2023evora}. Contrastive learning has shown promising results in learning expressive representations that can be used for traversability estimation in a self-supervised manner~\cite{Seo2023learning,Xue2023contrastive,Jung2023}.

\cite{Chen2023} generate pseudo labels from heuristic-based \lidar analysis, while~\cite{Frey2023} use a velocity-tracking criterion to assess terrain traversability to train a vision model. Both methods allow for training the traversability prediction network online during deployment.
In WayFAST~\cite{Gasparino2022}, the terrain traversability is approximated by the tracking error from a model predictive controller and used to predict traversability from image data. 
Similarly,~\cite{Cai2022} predict worst-case expected cost and traction from a semantic elevation map in addition to predicting confidence values using density estimation. 
The \textit{TerrainNet} framework~\cite{Meng2023} proposes using co-labeling of semantics by a human and a pre-trained semantic network, while the geometry is supervised by \lidar measurements. 

Our method uses self-supervision, but instead of predicting semantics or relying on heuristic-based analysis, we use the sophisticated field-tested traversability estimation software stack \gls{stack}, which allows us to account for a variety of risks in a probabilistic manner. 
This allows \glsentrylong{ours} to obtain a nearly dense supervision signal and account for uncertainty in the signal itself during learning, therefore overcoming problems associated with sparse labels when only considering the terrain vehicle interaction. Creating the supervision signal using hindsight imposes few constraints on specific motions to follow. This enables us to train on data collected by a human driver as well as on data collected during an autonomous mission. However, where and which training data to collect remain crucial design choices.

\subsubsection{Other Traversability Estimation Approaches}
Model-based approaches for assessing geometric traversability have attracted continuous research, particularly in the context of wheeled robots, where the wheel-to-terrain interactions can be more easily defined compared to more complex legged systems \cite{Wellhausen2023}. These approaches demonstrate good generalization, specifically when the underlying assumptions like the rigidity of the environment, hold. 
For this, recent works rely on analysis of point cloud data \cite{Fan2021,Cao2022,Xue2023}, mesh data \cite{Hudson2022}, or elevation maps \cite{Bouman2020,Fan2021,Wellhausen2023}.
On the other hand, data-driven approaches use simulation to collect data in a trial-and-error fashion to estimate traversability for wheeled \cite{Guzzi20} or legged robots \cite{Frey2022}.

Complementary, learned dynamics models from real-world data are capable of accounting for unmodeled effects in simulation and are used for planning. 
\cite{Kahn2021} and \cite{Kim2022learning} learn a forward dynamics model, where \cite{Kahn2021} predict future events and states of the robot from real-world data, and \cite{Kim2022learning} use a simulation to collect data.
\cite{Xiao2021} learn the inverse kinodynamics model of a wheeled robot from proprioception and \cite{Karnan2022} expand this concept by conditioning the model on visual data of future terrain patches to anticipate terrain interaction. Therefore, these approaches implicitly learn traversability information about the environment for the deployed robotic system. 

\gls{irl} aims to learn the underlying reward structure guiding an agent's behavior \cite{Wulfmeier2015maximum}. By using demonstration data and the \gls{irl} formulation, \cite{Wulfmeier2015maximum} train a neural network to predict cost maps for simple game-like environments. 
\cite{Triest2023} translated this concept to off-road autonomous driving and predicted vehicle-centric traversability grid maps.

\cite{Bouman2020} assess geometric traversability based on multi-fidelity terrain maps and superposition of individual traversability maps.
This work was expanded in \cite{Fan2021} and \cite{Fan2022}, where 
 traversability risk is modeled as a distribution allowing to account for different geometric risks while taking into account uncertainty. \cite{Dixit2023step} added further improvements for subterranean exploration and incorporated semantic risks assessed solely based on processing geometry. 
Our \glsentrylong{stack} software stack, used to generate the pseudo ground truth, adapts this perspective of traversability estimation and improves multiple components of the traversability assessment tailored to off-road driving. We explain the in-depth adaptation made to facilitate high-speed off-road navigation in \secref{sec:method-preliminary}.

\subsection{Learning Semantics in BEV Representation}
The mobility of ground robots is typically constrained by the terrain geometry. Under this reduced mobility assumption, it is sufficient to model the environment from a \gls{bev} perspective, in which the geometry, semantics, or traversability costs can be identified. 
One challenge is to associate features from a camera image into the \gls{bev} perspective. 
If no depth measurements are available, the inverse perspective mapping (IPM) algorithm under the flat ground assumption can project image features to the \gls{bev} perspective.
\cite{Roddick2019} showed that projecting multi-scale image features into 3D using an orthographic projection improves 3D object detection performance for autonomous driving.
This method projects features uniformly along a camera ray in 3D and consecutively pools the features along the z-dimension.
Similarly, \textit{Lift Splat Shoot}~\cite{Philion2022} extends this concept to multiple cameras and shows promising results for the task of semantic segmentation in the \gls{bev} perspective. Instead of densely distributing features along the ray, they predict a categorical distribution over discrete depth bins for each pixel indicating the contribution of each feature in the intermediate 3D representation.
Pseudo-\lidar approaches such as \textit{\mbox{BEV-Seg }}~\cite{Ng2022} predict explicit dense depth from the monocular image, which is used to establish a one-to-one correspondence between image pixels and grid cells.
\textit{Simple-BEV}~\cite{Harley2022} and \textit{M2BEV}~\cite{Xie2022} show that, while the feature lifting problem is well-researched, a uniform lifting approach~\cite{Roddick2019} can result in competitive performance and more gain can be achieved using RADAR sensor data~\cite{Harley2022} or a larger image backbone~\cite{Xie2022}. 
\textit{BEVFusion}~\cite{Liu2022} considers the problem of fusing \lidar data and camera images into a unified representation. Camera features are lifted into 3D following~\cite{Philion2022}, with a significantly more efficient implementation which addresses concerns by~\cite{Harley2022,Xie2022}. Depth features are concatenated in the \gls{bev} representation with image features, while no \lidar measurements are leveraged for the feature lifting. 
\textit{LAPTNet-FPN}~\cite{Diazzapata2023} use the \lidar measurements to lift camera features into 3D, by directly associating the sparse \lidar measurements to the downsampled image features. 
Alternative attention-based lifting approaches have been investigated in~\cite{Saha2022}.

\section{METHOD}
\label{sec:Method}
\begin{figure}[t]
    \centering
\includegraphics[width=1.0\columnwidth]{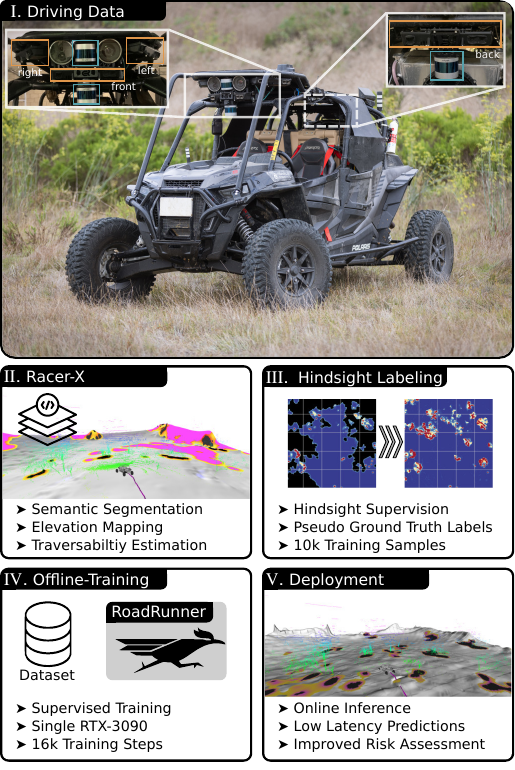}
    \caption{Overview of the \glsentrylong{ours} Architecture. The \glsentrylong{ours} network is trained on real-world driving data (I), which is first processed by \glsentrylong{stack} to generate an elevation and traversability assessment based on the currently available sensory data (II). Pseudo ground truth labels are generated by fusing information from past and future measurements to obtain reliable traversability and elevation estimates (III). The \glsentrylong{ours} network is trained offline based on the large dataset (IV) and can be deployed online for improved performance and reduced latency (V).}
    \label{fig:method-system_overview}
\end{figure}

\subsection{Problem Statement}
\label{sec:method-problem_statement}
Our \glsentrylong{ours} network predicts vehicle-centric elevation and traversability cost maps using multi-camera and multi-\lidar inputs.  
In the following sections, we describe the notation (\secref{sec:method-notation}) used throughout the paper and an overview of \glsentrylong{stack} stack (\secref{sec:method-preliminary}) used for generating training data in hindsight (\secref{sec:method-hindsight}). 
Next, we explain the network architecture (\secref{sec:method-road_runner_network}), followed by the implementation details (\secref{sec:method-implementation_details}). 
The overview of the RoadRunner architecture is presented in \figref{fig:method-system_overview}.

\subsection{Notation}
\label{sec:method-notation}

\begin{figure*}[t]
    \centering
\includegraphics[width=1.0\textwidth]{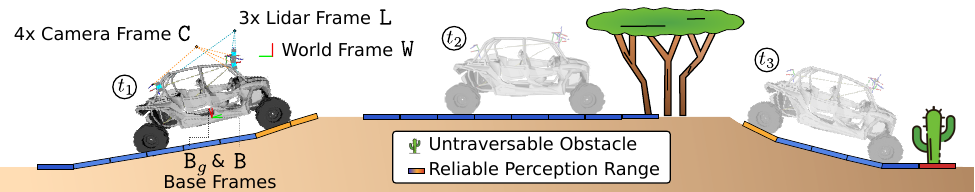}
    \caption{Definition of frames and illustration of hindsight self-supervision. At the first timestep, $t_1$, the reference frames are defined. The gravity-aligned base frame~$\frameBG{}$ is fixed to the vehicle (position and yaw), with roll and pitch being gravity-aligned. The tiled region below the vehicle on the ground illustrates the Reliable Perception Range per timestamp, where sufficient sensory information is available such that \glsentrylong{stack} can correctly predict the elevation and traversability (color of each tile). When the vehicle approaches the tree at timestamp $t_2$, \glsentrylong{stack} can correctly predict that the area underneath the canopy is traversable. Similarly, in timestamp $t_3$, the cactus can be identified as untraversable.
    While \glsentrylong{stack} requires exhaustive geometric information, which is only available in the proximity of the vehicle, more precise traversability and elevation maps, the so-called pseudo ground truth, can be generated as a learning objective for \glsentrylong{ours} when taking into account future and past sensory information. For example, in timestamp $t_2$, \glsentrylong{ours} can learn to correctly identify the cactus as a hazard from the image data, even with insufficient geometric information available. 
    }
    \label{fig:method-notation}
\end{figure*}

We use the following coordinate frames: world frame~$\frameW{}$, base frame~$\frameB{}$, gravity-compensated base frame~$\frameBG{}$, camera frames~$\frameC{k}$, and \lidar frames~$\frameL{l}$, with $k \in \{1, \dots , C\}$ and $l \in \{1, \dots, L\}$. Here, $C$ and $L$ denote the number of cameras and \lidar sensors, respectively.
The world frame~$\frameW{}$ in the following coincides with the typical odometry frame definition. Its relative transformation to the base frame~$\frameBG{}$ is assumed to be locally consistent and smooth. The frames are visualized in \figref{fig:method-notation}. 
Tensors are denoted by bold capital letters $\feat{A}{B}{C}{} \in \mathbb{R}^{A \times B \times C}$, where the lower right subscript indicates the respective dimensions. 
Camera images of height $H$ and width $W$ are denoted as $\img{k}{H \times W \times 3}{t}$, where $k$ indicates the respective camera index and $t$ is the timestamp.
The intrinsic calibration of the camera $k$ is given by $\K{k}$. 
Similarly, the point cloud measured by the \lidar $l$ is denoted as $\pcd{l}{N_l\times3}{t}$, where $N_l$ is the number of points and $t$ is the respective timestamp.

%
\begin{figure*}[p]
  \centering
  \includegraphics[width=0.85\textwidth]{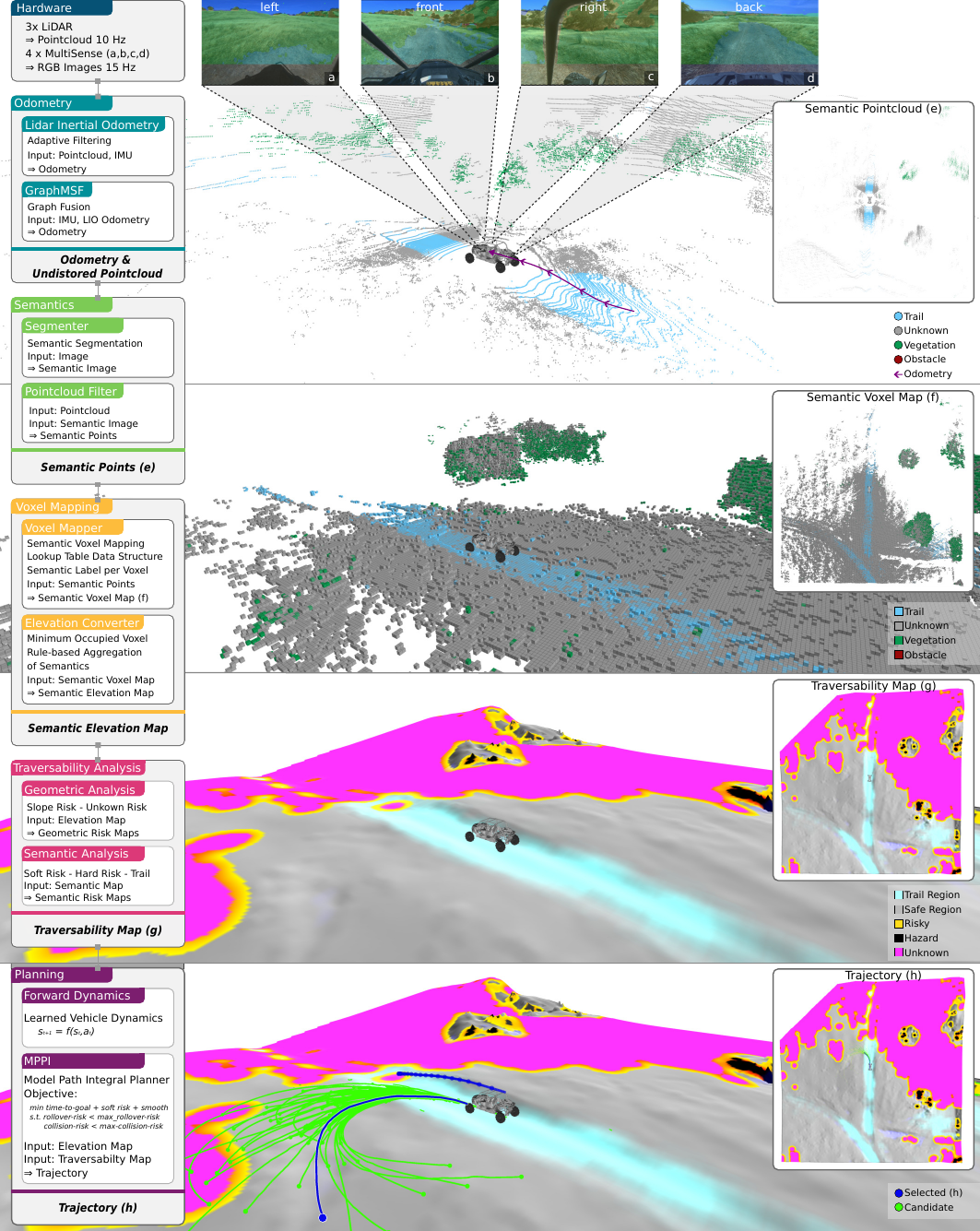}
  \caption{Overview of \glsentrylong{stack}. We use our \lidar Inertial Odometry system~\cite{Rose23} in combination with GraphMSF~\cite{Nubert22} to obtain smooth, accurate, and high-frequency odometry estimates. Segmenter~\cite{Strudel21} is used to predict the semantic classes from each camera image \textbf{\textit{(a--d)}}, which are then projected onto the undistorted and filtered point cloud, yielding the Semantics Points \textbf{\textit{(e)}}. The Semantic Points are further accumulated in a vehicle-centric voxel map using our voxel mapper based on~\cite{Overbye2021}, resulting in a Semantic Voxel Map. A rule-based aggregation method allows converting the Semantic Voxel Map further into a Semantic Elevation Map where the aggregation is tailored to off-road driving and the physical characteristics of our vehicle \textbf{\textit{(f)}}. Subsequently, both geometric and semantic risks are assessed based on the Semantic Elevation Map, resulting in a Traversability Map \cite{Fan2021} \textbf{\textit{(g)}}. For downstream trajectory optimization, both the traversability and the elevation are provided to a Model Predictive Path Integral (MPPI) Planner~\cite{Williams17}, which employs a learned vehicle dynamics model \cite{Gibson2023multistep} to compute the final trajectory for a given goal location \textbf{\textit{(h)}}.}
  \label{fig:method-stack_comparision}
  \vspace{-34.891pt}
\end{figure*}
We assume the point cloud scans are motion-compensated and merged for all $L$ LiDAR sensors, resulting in a single big point cloud $\pcdm{N \times 3}{t}$, with $N = \sum_{l=1}^{L} N_l$.
A grid map with dimension $H_g \times W_g$ is denoted as $\grid{H_g \times W_g}{}(x,y)$ and is expressed within the gravity-aligned base frame $\frameBG{}$. The frame $\frameBG{}$ defines the yaw orientation and center of the grid map, located at ($H_g/2$,$W_g/2$). Therefore we refer to the grid maps being position and orientation vehicle-centric.

\subsection{X-Racer Architecture}
\label{sec:method-preliminary}

\subsubsection{Overview} 
\glsentrylong{ours} leverages the highly capable mapping and traversability assessment \glsentrylong{stack} stack for learning. 
\glsentrylong{stack} allows the \glsentrylong{vehicle} to autonomously drive off-road at up to $\SI{15}{m/s}$ within a wide range of complex off-road environments. The task at hand, which guided the development of \glsentrylong{stack}, is to reach a set of goal positions using onboard sensing without reliance on precise prior global mapping in a GPS-denied environment. In the following, we will summarize the most important parts of \glsentrylong{stack} stack relevant to the traversability assessment used for local planning.
In addition, we will elaborate on the inherent limitations that are addressed by \glsentrylong{ours} with respect to the existing \glsentrylong{stack} stack. A simplified overview of all components can be seen in \figref{fig:method-stack_comparision}.

\subsubsection{Odometry and Semantics}  
We assume precise vehicle poses are provided by our \lidar-inertial odometry estimation and localization module~\cite{Fakoorian21} in combination with a graph-based filtering and fusion algorithm~\cite{Nubert22}.
Four RGB images captured by the \glsentrylong{camera} RGB-D cameras are individually processed by a Segmenter vision transformer model~\cite{Strudel21} adapted and fine-tuned using~\cite{mmseg2020}, to predict multi-class terrain semantics, which include classes such as trail, ground, vegetation, rock, and sky.
The segmentation network is trained in a supervised fashion using manually annotated images from real-world deployments and photo-realistic renderings using the Duality Robotics Falcon Pro simulator~\cite{Duality2023}. 

\subsubsection{Mapping} 
The point clouds sensed by three Velodyne VLP-32C \lidar sensors are filtered for vehicle self-correspondences using a predefined mask, and dust points are removed using an intensity-based filtering method. 
The filtered point cloud is associated with predicted semantics in the image domain based on proximity using the camera intrinsics and camera-to-\lidar extrinsic while correctly accounting for the ego-motion. 
The resulting semantic point cloud is probabilistically fused into a 3D vehicle-centric occupancy map covering a volume of $\SI{100}{m} \times \SI{100}{m} \times \SI{30}{m}$, which is defined as the \textit{micro range}, at an isotropic voxel resolution of $\SI{20}{cm}$.  
The 3D voxel mapper is a significantly enhanced version of \cite{Overbye2021}, which has been extended to allow for low-latency robot-centric mapping, voxel-decay, fusion of semantic segmentation, and a wide range of further features while being able to run in real-time on a single GPU. 
From the 3D occupancy map an elevation map is extracted, using the height of the lowest occupied voxel along the $z$-direction, followed by a sequence of geometric and semantics-based heuristics, to estimate the ground plane beneath vegetation and hazards.
A heuristic-based approach is used to fuse the semantic probabilities along the $z$-direction into the elevation map, which takes into account the distance of the voxel to the elevation map surface and its semantic class probabilities. 
E.g\onedot, all voxels up to a height of \SI{2.5}{m} above the elevation map labeled as a tree with a likelihood of over \SI{50}{\%} are fused into the elevation map cell. 
On the other hand, voxels above \SI{2.5}{m} labeled as \textit{tree} can be excluded to allow driving underneath the canopy. 

\subsubsection{Traversability Assessment} 
\label{subsub:method_trav_ass}
The traversability assessment is based on a previously developed planner for subterranean and unstructured environments~\cite{Fan2021}. It handles perceptual uncertainty based on the notation of certainty and risk. 
We provide an overview of the different risks taken into account to obtain the wheel risk, which describes the risk associated with an individual wheel interacting with the terrain.
First, a reliability map is computed, which is proportional to the density of the available geometric information. For example, if insufficient geometric observations about a specific area within the elevation map are available, the estimates for these cells are considered to be less reliable. 
Further, the elevation map is interpolated and smoothed such that the risk for slope, curvature, roughness, and positive and negative obstacles can be extracted. Additional semantic risks for hard and soft obstacles are computed. The final wheel risk, which describes the risk associated with the terrain-wheel interaction, is given by the Conditional Value At Risk (\gls{cvar}) of the combined assessed risk distributions. 
The full \glsentrylong{stack} software stack is tuned in simulation and on real-world data. 
In the following, we refer to the CVaR of the wheel risk as the traversability, where a value of $0$ indicates safe to traverse, whereas $1$ is unsafe.

\subsubsection{Trajectory Planning} 
The resulting \gls{cvar} wheel risk traversability map of size $\SI{100}{m} \times \SI{100}{m}$ at a resolution of $\SI{20}{cm}$ is provided to the \gls{mppi} planner \cite{Williams2018} responsible for obstacle avoidance at high speed. The planner operates with a \SI{5}{s} planning horizon and is evaluated at a frequency of \SI{30}{Hz}.
The planner has access to the dynamics model of the vehicle and minimizes the time to the goal while trading off soft risks and smoothness of the path. The planned trajectories have to obey rollover, collision, and velocity limitations strictly. 

\subsubsection{Limitations}
While the current software stack is capable of safely guiding the vehicle within desert environments and individual components can be tuned, multiple limitations exist, which are addressed by \glsentrylong{ours}. 
Geometric perception of the environment is the dominant driver of \glsentrylong{stack}: semantic information is only incorporated if geometric measurements are available. 
Therefore, a risk identified only in the image domain cannot be taken into account.
This problem is specifically pronounced when operating at higher speed, given the limited update rate of the \lidar (\SI{10}{Hz}) and its sparse nature at long distances. 
Additionally, while the set of considered semantic classes is carefully selected, it inherently limits the flexibility of the downstream traversability assessment to account for semantic risks. 
Further, while the sequential structure of the stack allows for interpretability and adjustment of individual components, it leads to a high latency between perception and traversability estimation of over \SI{500}{ms}. This high latency limits the maximum speed, which is determined by the emergency braking distance and hazard detection range. 
Lastly, the current software stack cannot make use of (past) experience, in contrast to a rally driver who can anticipate the road progression when turning into a narrow corner and thus take more sophisticated actions under uncertainty. \glsentrylong{stack} stack does not include these generative abilities to imagine the environment, which we argue are important for driving at high speeds.

\subsection{RoadRunner Hindsight Traversability Generation}
\label{sec:method-hindsight}
Using the \glsentrylong{stack} software stack we generate training data for \glsentrylong{ours} in a self-supervised manner using hindsight. In particular, the traversability and elevation are improved by leveraging privileged future measurements, not available to \glsentrylong{stack} during run-time. This reduces the uncertainty about the environment and allows for the extension of the reliable prediction range beyond the current horizon. An illustrative example is presented in \figref{fig:method-notation}. 
This is implemented by accumulating predictions obtained by \glsentrylong{stack} over time, requiring no modifications of major components such as the mapping backend. 
To reduce errors due to localization drift, all transformations are performed in the locally consistent odometry frame. Only measurements collected within a time window of $\SI{60}{s}$ are accumulated for each grid map. 

\RestyleAlgo{ruled}
\begin{algorithm}
\SetAlgoNlRelativeSize{0}
\SetAlgoNlRelativeSize{-1}
\SetAlgoNlRelativeSize{-2}
\newcommand\graycomment[1]{\footnotesize\ttfamily\textcolor{gray}{#1}}
\SetCommentSty{graycomment}
\SetKwFunction{Sort}{Sort}
\SetKwFunction{Transform}{Transform}
\SetKwFunction{Fusion}{Fusion}
\SetKwInOut{Input}{Input}
\SetKwInOut{Output}{Output}

\BlankLine  
  $G\_list \; [\;] \leftarrow$ \text{grid maps within 60s window}\; 

  $p \leftarrow$ \text{reference $SE(2)$ position}\; 

  $default \leftarrow$ \text{default value ground truth map}\; 
  
  \Fusion $\leftarrow$ \text{fusion function}\; 

  \BlankLine

\tcp{Sort with increasing timestamp}
\Sort{$G\_list$}\;

\tcp{Initalize the return grid map}
$G\_gt \leftarrow \text{torch.full}(G\_list[0].\text{shape}, default)$\;

\BlankLine
\ForEach{$G$ \textbf{in} $G\_list$}{
  \tcp{Transform the grid map to G\_gt frame}
  $G\_transformed \leftarrow \Transform(G, p)$\;
  
  \tcp{Fuse information into G\_gt}
  $G\_gt \leftarrow \Fusion(G\_gt, G\_transformed)$\;
}

\BlankLine
\Return{$G\_gt$}\;

\caption{Compute Hindsight Ground Truth}
\label{alg:compute-hindsight}
\end{algorithm}

Algorithm \ref{alg:compute-hindsight} describes the fusion scheme used for the pseudo ground truth reliability, elevation, and traversability map generation. 
In the following, we will designate this as the ground truth for simplicity, omitting the term "pseudo". It is important to note, however, that the presented ground truth does not reflect the actual reliability, elevation, or traversability, but can be used as a good approximation.

The default value for each map is initialized to be unreliable (value $0$), the maximum height ($+\infty$), and traversable (value $0$), respectively.
Each predicted grid map within the given time window is first translated and rotated to the reference position. 
The fusion function is given by the cell-wise minimum for the elevation, the cell-wise maximum for the reliability, and the temporally latest measurement in addition to a confidence threshold for the traversability. 
The traversability fusion is deliberately not chosen to be a simple maximum function, which one could argue facilitates safety. Experimentally, the maximum function results in an over-pessimistic traversability assessment given that under uncertainty areas may be predicted as untraversable by \glsentrylong{stack}. Later, during deployment, with more information available, these areas can often be classified as safe to traverse.

\subsection{RoadRunner Network}
\label{sec:method-road_runner_network}
RoadRunner is specifically designed to allow for high-speed off-road navigation. This requires RoadRunner to operate at a low latency of~\SI{140}{ms}, such that a new prediction of the environment can be provided every \SI{3}{m}, when driving at a speed of up to \SI{20}{m/s}. 
Therefore, we choose a prediction range of $\SI{100}{m} \times \SI{100}{m}$, which is adequate for handling high speeds. 
We also directly operate on single LiDAR scans rather than a fused map representation. This makes RoadRunner less susceptible to sparse geometric information at high speeds. The number of geometric points per LiDAR point cloud is nearly independent of the vehicle velocity. This is not the case when operating on an accumulated map representation, where the density of geometric information strongly correlates with the vehicle velocity. 

\subsubsection{Preprocessing and Feature Extraction}
\begin{figure*}[t]
    \centering
    \includegraphics[width=\textwidth]{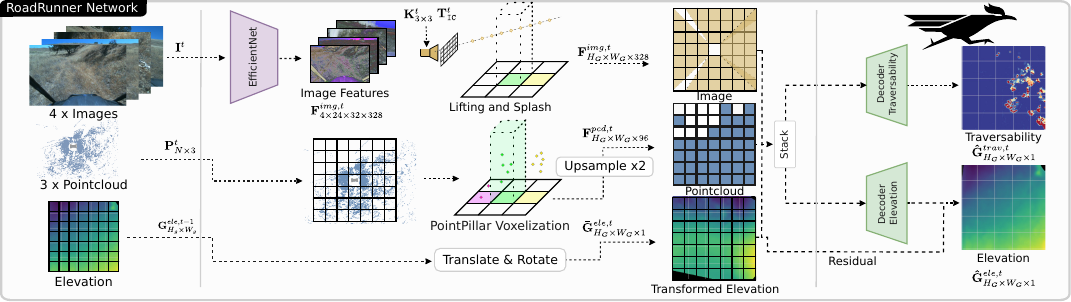}
    \caption{Overview of the \glsentrylong{ours} Network. The input to the network consists of 4 RGB images and a filtered and merged point cloud from 3 \lidar sensors, in addition to the past elevation prediction of the previous timestamp $t-1$. 
    The network uses the \textit{Lift Splat Shoot} \cite{Philion2022}, and the PointPillar \cite{Lang2019} architecture to encode the visual and geometric information, respectively. The elevation information is normalized and transformed to the current position, which is then used to predict the traversability and elevation based on separate decoder networks.}
    \label{fig:method-network_overview}
\end{figure*}

The \glsentrylong{ours} network builds upon the open-source available implementation of \textit{Lift Splat Shoot}~\cite{Philion2022}, \textit{PointPillars}~\cite{Lang2019} and \textit{BEVFusion}~\cite{Liu2022}, which combines and improves the prior two methods. 
The information flow and \glsentrylong{ours} network architecture is shown in \figref{fig:method-network_overview}.
The images obtained from the hardware-synchronized cameras are first downsampled to a resolution of $512\times384$ and then normalized. The image timestamp $t$ determines the reference frame of the vehicle-centric grid map. 
The latest measurement of each \lidar is motion compensated, transformed to the base reference frame $\frameBG{}$, and merged to a single merged point cloud $\pcdm{N\times3}{t}$. 
Features are extracted from each camera image using an EfficientNet-B0 \cite{Tan19}, which downsamples the input size by a factor of $16$, yielding a corresponding feature map $\feat{32}{24}{328}{k}$ for each image~$k$. 
The merged point cloud is passed through a \textit{PointPillars} backbone with the same grid configuration as the target grid map of $\SI{100}{m} \times \SI{100}{m}$ at a resolution of $\SI{20}{cm}$ and results in a point cloud feature map $\feat{H_G/2}{W_G/2}{96}{pcd}$. 
The previous elevation estimate $\mathbf{G}_{H_G \times H_W \times 1}^{ele, t-1}$ is translated and rotated to align with the new vehicle position. In addition, the elevation is normalized by first scaling the elevation by a factor of $0.05$ and then clipping to values between $-1$ and $1$, resulting in $\mathbf{\bar{G}}_{H_G \times H_W \times 1}^{ele, t}$. This scaling limits the range of elevation to $\pm \SI{20}{m}$, which is adequate for our off-road scenario (refer to the normalized elevation histogram in~\figref{fig:method-normalization}). 

\subsubsection{Lifting}
The lifting of a feature from the camera image plane into the 3D space is achieved by using the pinhole camera model, following \cite{Philion2022}. Along the respective ray of the pixel, a set of equally spaced discrete points in 3D are distributed, resulting in a set of feature points.
The weight of a feature point is given by a categorical distribution predicted along the ray based on the image feature embedding of the respective pixel using a $1 \times 1$ convolutional layer. 
The feature points are rasterized into grid map cells with the same dimensions and resolution as the target grid map.
The weighted feature points coinciding with a grid cell are accumulated by a channel-wise summation using the improved \textit{splat} implementation proposed in \cite{Liu2022}, resulting in the image feature map $\feat{H_G}{W_G}{328}{img}$.

Next, we emphasize why we express the grid map in the gravity-aligned base frame $\frameBG$. This choice offers a primary advantage of accumulating information along the gravity-aligned $z$-direction when performing the \textit{splat} operation. In particular, when considering driving up a steep incline, the information about vertical obstacles such as a tree trunks or vertical poles is accumulated within a single grid cell, instead of washed out over multiple cells.
If no feature point is associated with a grid cell, it is set to $0$. 

\subsubsection{Multi-Modal Fusion}
First, the point cloud feature map is upsampled to a resolution of $H_G \times W_G \times 96$. The three feature maps, namely  
$\feat{H_G}{W_G}{328}{img}$, $\feat{H_G}{W_G}{96}{pcd}$, and $\mathbf{\bar{G}}_{H_G \times H_W \times 1}^{ele, t}$ are concatenated resulting in the multi-modal feature map of size $\feat{H_G}{W_G}{425}{fused}$. Supplying the nearest neighbor interpolated elevation map to the network facilitates the preservation of elevation information for regions that are unobservable within the currently processed merged LiDAR point cloud. 
Two separate decoder networks are trained for elevation and traversability estimation, predicting $\gridhat{H_G \times W_G}{ele}$ and $\gridhat{H_G \times W_G}{trav}$, respectively. 
Each decoder consists of a set of convolutional, batch norm, and ReLU activation layers following a residual architecture. 
The elevation decoder predicts the residual elevation to the motion-compensated elevation. To enforce a smooth traversability output, a median filter of window size 5 $\times$ 5 is applied to the predicted elevation. 

\begin{figure}[t]
    \centering
    \includegraphics[width=0.5\textwidth]{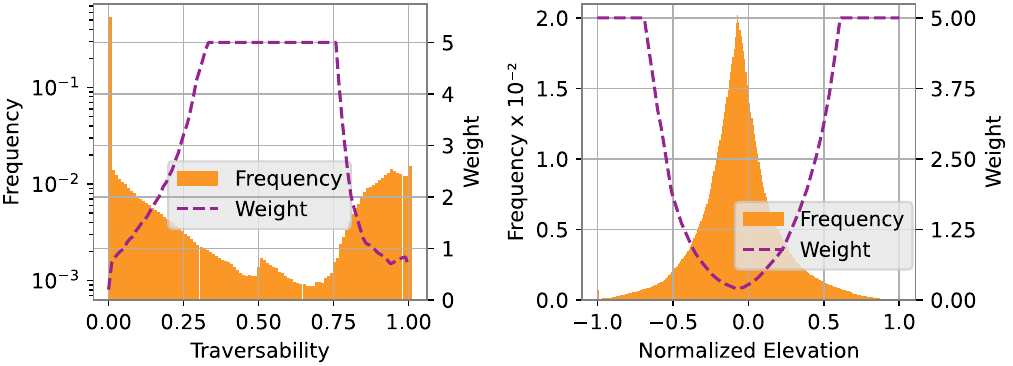}
    \caption{Normalization weights for the traversability and normalized elevation. The left axis provides the frequency per bin. The right axis (magenta dotted line) provides the weight. The wheel risk is strongly imbalanced, with most areas being fully traversable (value of 0). The elevation follows a mirrored exponential distribution, with its peak being slightly negative, given that the base frame $\frameBG$ is above the ground.}
    \label{fig:method-normalization}
\end{figure}

\subsubsection{RoadRunner Loss}
\label{subsubsec:loss}
To train \glsentrylong{ours}, we use the \gls{wmse} for the traversability and elevation. The loss is calculated only for the valid grid cells containing a supervision signal: 
\begin{equation}\mathcal{L}_{\mathrm{WMSE}}^{\text{layer}} = \frac{1}{\lvert \grid{}{} \rvert} \sum_{x,y \, \in \, \grid{}{}}w^{\text{layer}}(\grid{}{}(x,y) - \gridhat{}{}(x,y))^2.
\end{equation}
Here, the weighting factor $w^{\text{layer}}$ is calculated based on the value of the target grid cell. 
For this, we compute the normalized frequency $\mathcal{F}_{N_{\text{bins}}}$ of elevation and traversability scores over the training dataset in $N_{\text{bins}}$ bins. The clipped inverse normalized frequency gives the final weight:
\begin{equation}
    w^{\text{layer}} =  \text{clip}( (\mathcal{\bar{F}}^{\text{layer}} * N_{\text{bins}})^{-1}, 0.2, 5).
\end{equation}
Multiplying by the number of bins results in a weight of $1$ for all bins with a uniform probability.
The relation between normalized frequency and resulting weight is visualized in \figref{fig:method-normalization}.
The final optimization objective is given by: 
\begin{equation}
    \mathcal{L}_{\mathrm{FINAL}} = \mathcal{L}_{\mathrm{WMSE}}^{\mathtt{trav}} + \mathcal{L}_{\mathrm{WMSE}}^{\mathtt{elev}}.
\end{equation}

\subsection{Implementation Details}
\label{sec:method-implementation_details}
We use the AdamW optimizer \cite{Ilya19} with a learning rate of $10^{-3}$, OneCycleLearningRate-schedule \cite{Leslie17}, and optimize for a total of $16,000$ steps. The network is trained on an Nvidia RTX3090 GPU with a batch size of $6$. For the EfficientNet-B0 we use pre-trained weights from ImageNet, and freeze the respective layers. For the lifting operation, we generated a total of $353,280$ feature points given by the downsampled height of $24$, downsampled width of $32$, $4$ cameras, and $N_D=230$ feature points, where we selected the closest distance being \SI{4}{m} and the furthest distance being \SI{50}{m} with a spacing of \SI{0.2}{m} analog to the grid resolution.
In total, \glsentrylong{ours} accounts for \SI{24.0}{M} parameters, with \SI{5.3}{M} EfficientNet-B0, \SI{4.4}{M} \textit{PointPillars}, \SI{4.6}{M} \textit{Lift Splat Shoot}, and the remaining \SI{9.8}{M} parameters correspond to the decoder networks.


\section{EXPERIMENTS}
\label{sec:Experiments}
\begin{figure*}[t]
    \centering
    \includegraphics[width=\textwidth]{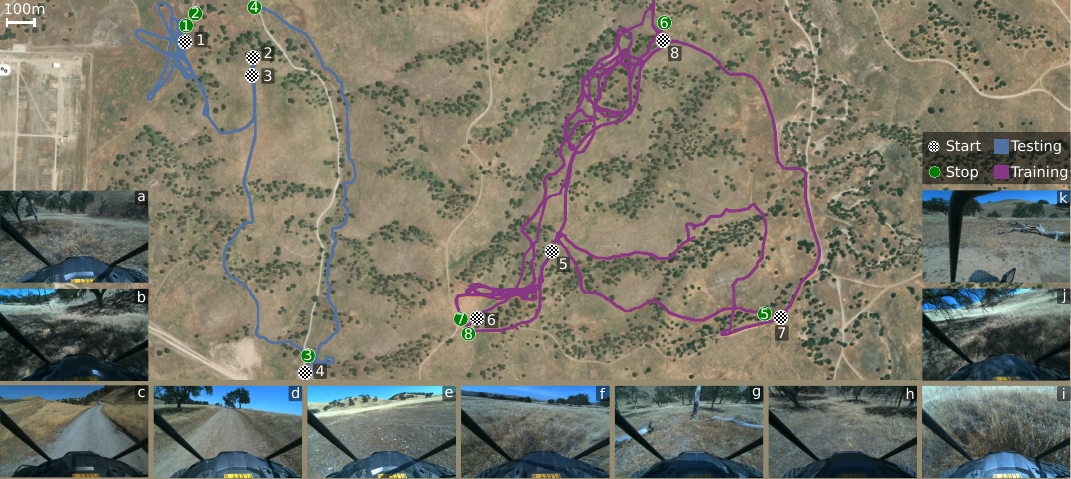}
    \caption{Dataset Overview: \textcolor{magenta_dark}{Magenta trajectories} are used for training and validation. \textcolor{blue}{Blue trajectories} are used for testing. The dataset comprises various driving scenarios illustrated in \figref{fig:experiments-dataset}, including open-field \textbf{\textit{(e)}}, dirt road \textbf{\textit{(d)}}, and gravel road traversal \textbf{\textit{(c)}}, as well as operation within confined canyon/ditch settings \textbf{\textit{(a)}} and forested environments \textbf{\textit{(a, b, h)}}. Notably, the forest environment posed unique challenges, necessitating driving beneath the canopy \textbf{\textit{(b, j)}}. Varying lighting conditions including shadows from trees \textbf{\textit{(a, b, j)}}, substantially impeded vision-based obstacle detection. Additionally, varying levels of vegetation \textbf{\textit{(e, f)}} and the presence of fallen trees \textbf{\textit{(g, j)}} within tall grass poses additional challenges for obstacle detection. 
    }
    \label{fig:experiments-dataset}
\end{figure*}

\subsection{Dataset - Scenarios}
To train our model, we collect a total of \SI{16.5}{km} of off-road driving data at Paso Robles, CA, USA. The dataset consists of eight individual trajectories resulting in a total of $21,158$ samples.
We apply our hindsight traversability generation (\secref{sec:method-hindsight}) on all collected trajectories. 
The trajectories are split into training, validation, and testing data (see.~\figref{fig:experiments-dataset}). 
The test data is collected in a different geographic location compared to the training and validation data. However, the test data is from the same ecoregion with similar topography and vegetation. For training and validation, four trajectories are collected and roughly the first $80\%$ are used for training while the last $20\%$ are used for validation. 
\tabref{tab:experiments-dataset} summarizes the key statistics about the dataset.

\begin{table}[hb]
    \centering
    \ra{1.2}
    \setlength{\tabcolsep}{3pt}
    \caption{Dataset Overview - Statistics recorded in Paso Robles}
    \begin{threeparttable}
    \begin{tabular}{lcrcc}
    \toprule
    Split & Duration       & Distance         & \# Traj  & \# Samples  \\ \midrule 
    Train & \SI{3389}{s}   & \SI{8618}{m}     & $4^+$    & 10626       \\
    Val   & \SI{644}{s}    & \SI{2236}{m}     & $4^-$    & 2658        \\
    Test  & \SI{1700}{s}   & \SI{5727}{m}     & 4        & 7874        \\ \bottomrule
    \end{tabular}
      \begin{tablenotes}
        \item[$+/-$ ] Indicates the first $80\%$ ($^+$) or last $20\%$ ($^-$) of the same trajectory respectively. 
      \end{tablenotes}
    \end{threeparttable}
    \label{tab:experiments-dataset}
\end{table}

Each recorded trajectory is converted into a set of samples by selecting camera images based on a minimal traveled distance criterion between samples of \SI{20}{cm}. In practice, this maintains the diversity of the data while reducing storage and computing requirements during training.
Each sample consists of the merged \lidar scan obtained by all LiDARs, $4$ camera images, elevation map, and the ground truth elevation and ground truth traversability map.

\subsection{Metrics}
We assess the \textit{elevation mapping performance} by reporting the \gls{mae}, following~\cite{Meng2023}: 
\begin{equation}
    \mathcal{L}_{\mathrm{MAE}} = \frac{1}{\lvert \grid{}{} \rvert} \sum_{x,y \, \in \, \grid{}{}} \lvert \grid{}{}(x,y) - \gridhat{}{}(x,y)  \rvert,
\end{equation}
where $x,y \, \in \, \grid{}{}$ consists of all valid indices of target grid map layer $\grid{}{}$, with $\gridhat{}{}$ denoting the estimated grid map layer. 
For the \textit{traversability estimation performance} we follow \cite{Frey2023} and evaluate the \gls{mse}:
\begin{equation}
    \mathcal{L}_{\mathrm{MSE}} = \frac{1}{\lvert \grid{}{} \rvert} \sum_{x,y \, \in \, \grid{}{}} (\grid{}{}(x,y) - \gridhat{}{}(x,y))^2.
\end{equation}
In addition, we introduce the reliability-weighted performance measures. The \gls{wmae} and \glsentrylong{wmse}~(\gls{wmse}) are given as:
\begin{equation}
    \mathcal{L}_{\mathrm{WMAE}} = \frac{1}{\lvert \grid{}{} \rvert} \sum_{x,y \, \in \, \grid{}{}}\mathbf{C}(x,y) \lvert \grid{}{}(x,y) - \gridhat{}{}(x,y)  \rvert,
\end{equation} 
\begin{equation}
    \mathcal{L}_{\mathrm{WMSE}} = \frac{1}{\lvert \grid{}{} \rvert} \sum_{x,y \, \in \, \grid{}{}}\mathbf{C}(x,y) (\grid{}{}(x,y) - \gridhat{}{}(x,y))^2,
\end{equation}

where $\mathbf{C}(x,y)$ denotes the reliability per grid cell and is directly estimated by \glsentrylong{stack} (\secref{subsub:method_trav_ass}). 

While the previously introduced metrics (\gls{mse}, \gls{wmse}) evaluate traversability as a continuous-valued function, it is also crucial to classify hazardous regions. This is necessary as downstream planning modules explicitly account for those regions. The distance at which hazard regions can be identified reliably bounds the maximum safe velocity of the vehicle, considering both the emergency braking distance and latency. 
Similarly to \glsentrylong{stack}, we introduce a \textit{fatal risk value} threshold. This threshold enables the conversion of the continuous predicted traversability risk map to a binary classification output. We use the same threshold as applied to \glsentrylong{stack} and utilize it for both the ground truth traversability map and the predicted traversability by \glsentrylong{ours}. This approach allows us to evaluate Precision, Recall, and F1-Score in terms of hazard detection (HD).

\subsection{Traversability Estimation Performance}

The primary focus is to investigate the performance of \glsentrylong{ours} in comparison to \glsentrylong{stack}. We use the nearest neighbor interpolation to fill the missing traversability estimates of \glsentrylong{stack}.
\glsentrylong{ours} outperforms \glsentrylong{stack} in terms of traversability prediction in MSE ($0.0113$ to $0.0216$), Precision ($0.412$ to $0.217$), while slightly underperforming in Recall for the selected \textit{fatal risk value} (cf.~\tabref{tab:experiment_perf_trav}). 
The numerical value of the traversability prediction is strongly dependent on the dataset (compare the difference between the validation and test datasets). This is due to the different environments and vehicle motion. While, e.g., driving down a hill, sufficient information about the environment is available for RoadRunner to output reliable predictions when driving up a hill and approaching the summit, no information about the summit may be available due to occlusions rendering this data inherently difficult, both for RoadRunner and for \glsentrylong{stack}. Despite the numerical difference between the training and test data, RoadRunner’s relative performance is consistent compared to \glsentrylong{stack} on the validation and test data.

\begin{figure*}[t]
    \centering
        \begin{minipage}[p]{0.5\linewidth}
        \centering
        \includegraphics[width=1\linewidth]{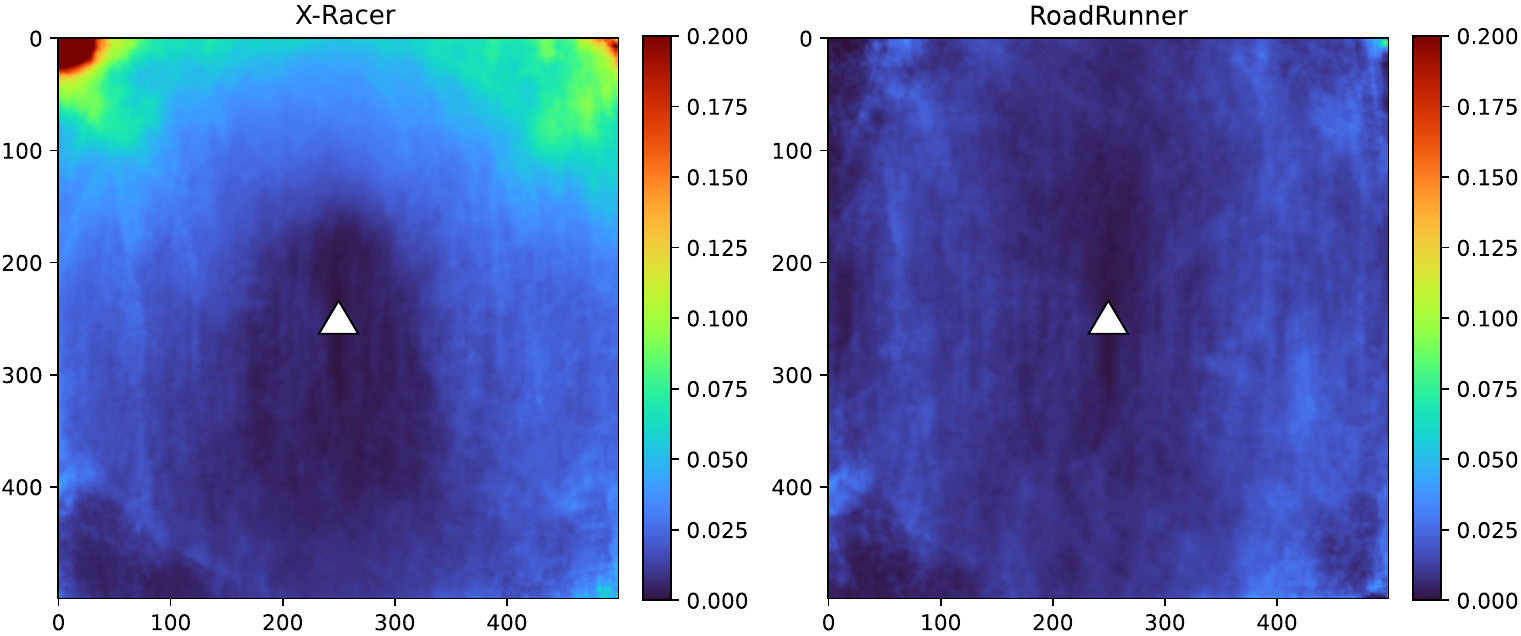}
        \subcaption{Traversability Prediction in MSE.}
        \label{fig:experiments-error_maps_traversability}
        \end{minipage}
        \begin{minipage}[p]{0.48\linewidth}
        \centering
        \includegraphics[width=1\linewidth]{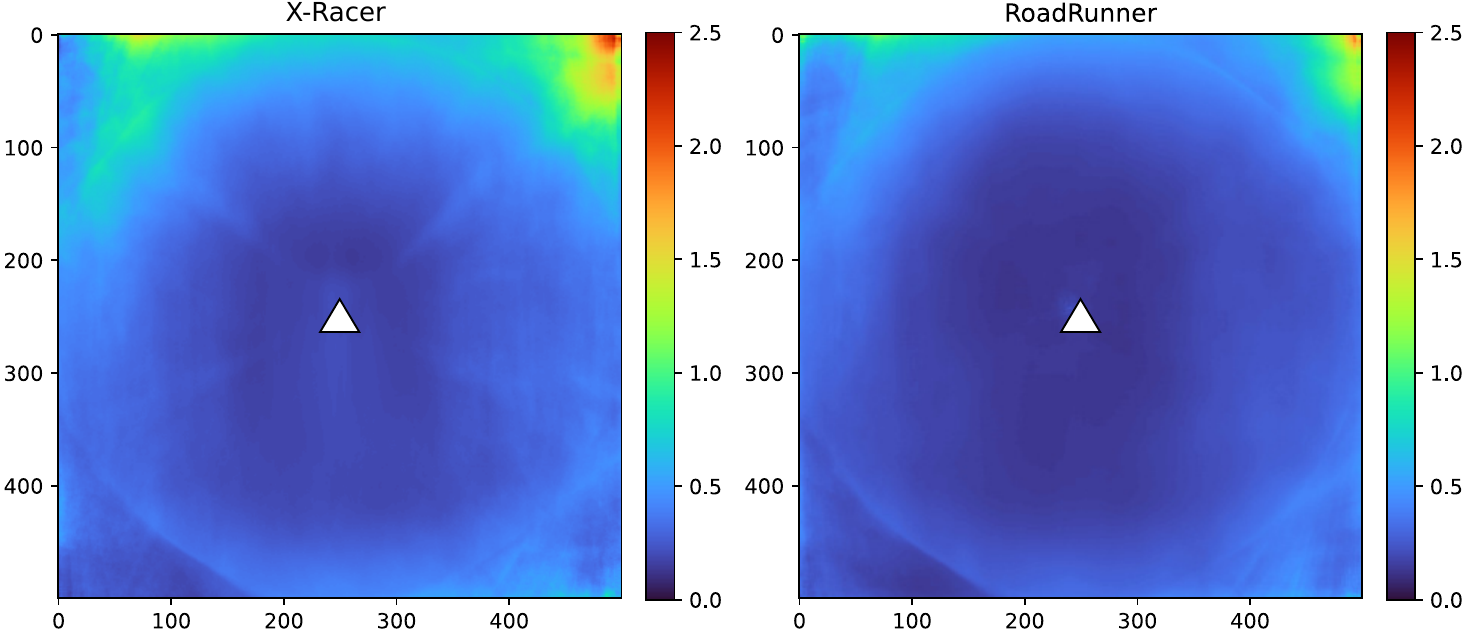}
        \subcaption{Elevation Prediction in MAE.} 
        \label{fig:experiments-error_maps_elevation}
        \end{minipage}
    \caption{Qualitative comparison of \glsentrylong{stack} and \glsentrylong{ours} per grid cell for traversability prediction \textbf{\textit{(a)}} and elevation prediction \textbf{\textit{(b)}} on average across all samples within the test dataset. The vehicle is located at the center of the map indicated by the $\triangle$, and the front of the vehicle faces towards the top of the grid map. \glsentrylong{ours} achieves an overall lower elevation and traversability prediction error. \glsentrylong{stack} performs specifically worse in front of the vehicle, where insufficient geometric data has been observed. }
    \label{fig:experiments-error_maps}
\end{figure*}

\begin{table*}[ht]
	\centering
	\ra{1.2}
	\setlength{\tabcolsep}{3pt}
 	\caption{Evaluation of Traversability Estimation Performance}
    \begin{tabular}{lcclcclcclcclcc} \toprule
        & \multicolumn{2}{c}{MSE$\downarrow$} && \multicolumn{2}{c}{WMSE$\downarrow$}  && \multicolumn{2}{c}{HD-Recall$\uparrow$} && \multicolumn{2}{c}{HD-Precision$\uparrow$}\\
        \cmidrule{2-3} \cmidrule{5-6} \cmidrule{8-9} \cmidrule{11-12} 
       Method & \glsentrylong{stack} & (ours)  && \glsentrylong{stack} & (ours) && \glsentrylong{stack} & (ours)  && \glsentrylong{stack} & (ours)\\
        \midrule 
        Train & 0.0339 & \textbf{0.0066}  &&  0.0339 & \textbf{0.0066} && 0.292 & \textbf{0.896} && 0.344 & \textbf{0.777} \\ 
        Val   & 0.0413 & \textbf{0.0295}  &&  0.0413 & \textbf{0.0295} && 0.305 & \textbf{0.364} && 0.399 & \textbf{0.512} \\ 
        Test  & 0.0216 & \textbf{0.0113}  &&  0.0216 & \textbf{0.0113} && \textbf{0.320} & 0.302 && 0.217	& \textbf{0.412} \\ 
    \bottomrule
    \end{tabular}
\label{tab:experiment_perf_trav}
\end{table*}
The MSE per grid cell in the vehicle-centric frame averaged across all samples within the test dataset is depicted in \figref{fig:experiments-error_maps}.
As the vehicle is moving in a straight line, it corresponds to an upward movement in the grid map. 
Given that the test data exhibits a bias towards driving straight, we expect a better performance in the lower part of the grid map, given that on average more information is accumulated behind the vehicle. 
The MSE distribution of \glsentrylong{stack}, directly confirms and motivates the rationale behind our proposed hindsight ground truth generation. This stems from the observation that lower MSE is achieved in the lower part of the grid map, which has already been passed by the vehicle. Additional information can be gathered and accumulated for those regions within the semantic voxel map, resulting in an improved traversability estimate. On the other hand, in front of the vehicle, insufficient geometric information is available and the nearest neighbor interpolation leads to a higher error. These results prompt our hypothesis that \glsentrylong{ours} can more effectively make use of the information provided within the three front-facing cameras and, therefore, estimate traversability better than \glsentrylong{stack}.

\begin{figure*}[t]
    \centering
    \includegraphics[width=\textwidth]{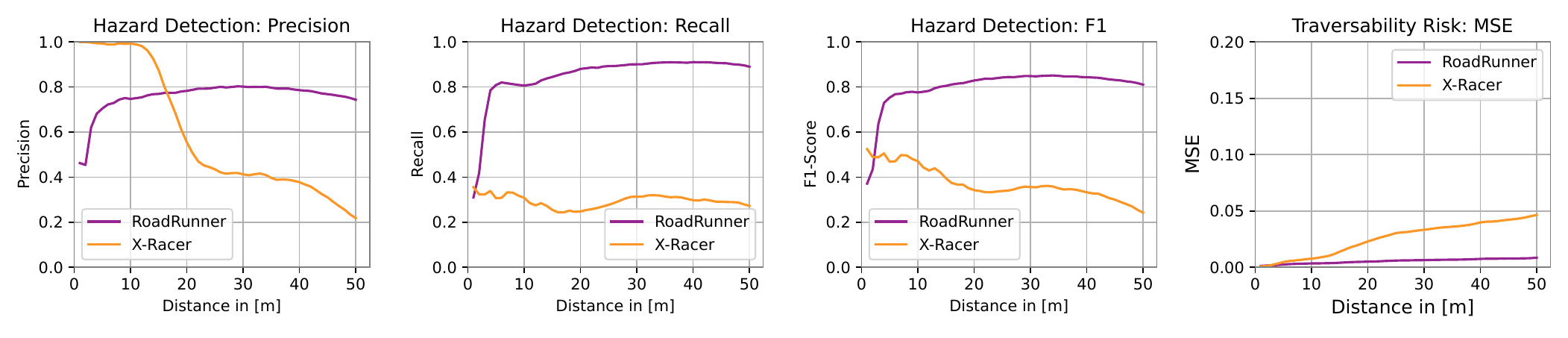}
    \includegraphics[width=\textwidth]{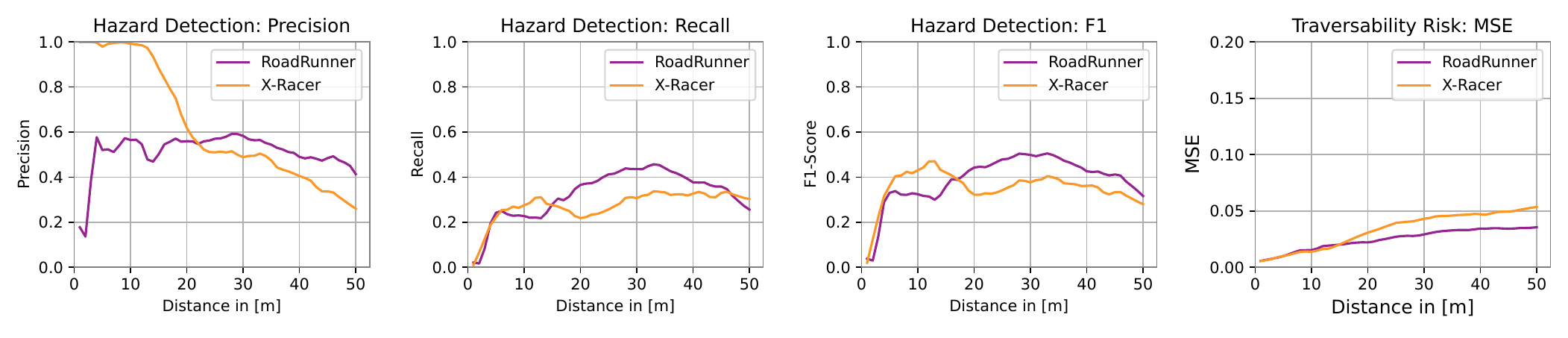}
    \includegraphics[width=\textwidth]{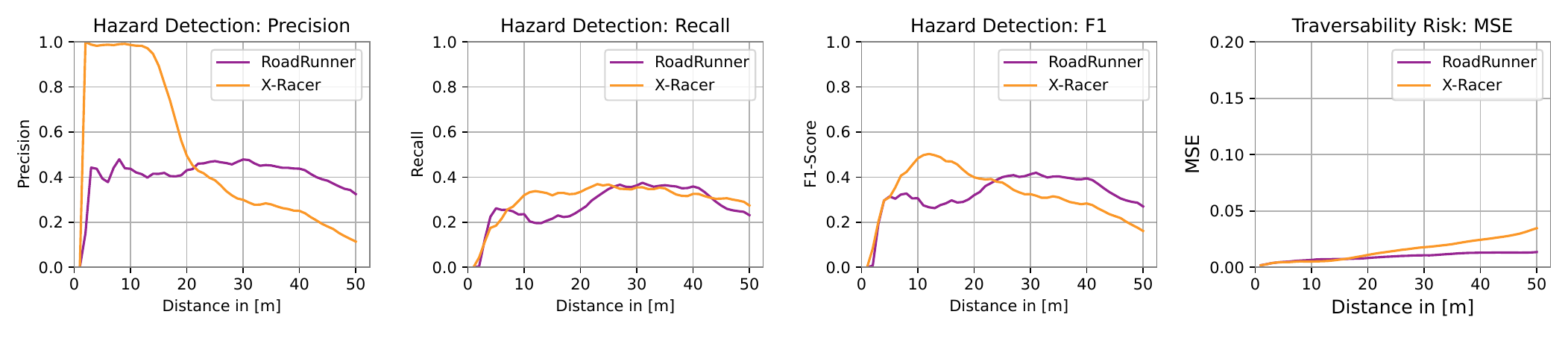}
    \caption{Hazard Detection Error Metrics of \glsentrylong{ours} and \glsentrylong{stack} on the training \textbf{(1st row)}, validation \textbf{(2nd row)} and test \textbf{(3th row)} dataset. We report the Precision \textbf{(1st column)}, Recall \textbf{(2nd column)}, F1-Score \textbf{(3rd column)} and the MSE \textbf{(4th column)}, plotted over the distance. We report the error evaluated for each 1-meter bin independently as opposed to the cumulative error up to the distance. 
    \glsentrylong{ours} outperforms \glsentrylong{stack} across all dataset above a range of \SI{20}{m} in terms of MSE. In the proximity to the vehicle, \glsentrylong{stack}, as expected, corresponds nearly perfectly to the ground truth resulting in perfect predictions.}
    \label{fig:experiments-wheel-risk-distance}
\end{figure*}

The hazard detection performance plotted over the distance is depicted in \figref{fig:experiments-wheel-risk-distance}. 
Fig. 9 depicts the hazard detection performance plotted over the distance. We highlight that at close range,\glsentrylong{stack} approximately corresponds to the ground truth; therefore, we do not expect \glsentrylong{ours} to outperform \glsentrylong{stack}. As expected in the vehicle's vicinity, \glsentrylong{stack} detects all hazards correctly (Precision of 1.0); however, at longer ranges, \glsentrylong{ours} outperforms \glsentrylong{stack}. Above a distance of \SI{25}{m}, \glsentrylong{ours} achieves a higher F1 score than \glsentrylong{stack} while consistently outperforming \glsentrylong{stack} in terms of MSE.

\subsection{Elevation Mapping Performance}
We report the elevation map performance between \glsentrylong{ours} and \glsentrylong{stack} in \tabref{tab:experiment_perf_elevation}.
Before interpreting the results, we would like to recapitulate the generation of the elevation map procedure by \glsentrylong{stack}. \glsentrylong{stack} fuses geometric information in a structured approach into a volumetric map, which is subsequently used in \secref{sec:method-hindsight} to compute the ground truth elevation. 
Consequently, when comparing the \glsentrylong{stack} against the ground truth, only a substantial performance increase can be expected for unobserved and therefore interpolated regions where no geometric measurements are available for the elevation mapping performance.
\begin{table}[ht!]
	\centering
	\ra{1.2}
	\setlength{\tabcolsep}{2.4pt}
 	\caption{Evaluation of Elevation Mapping Performance}
    \begin{tabular}{lcclcc} \toprule
        & \multicolumn{2}{c}{MAE [m] $\downarrow$} && \multicolumn{2}{c}{WMAE [m] $\downarrow$}  \\
        \cmidrule{2-3} \cmidrule{5-6}
        & \glsentrylong{stack}& (ours) && \glsentrylong{stack}& (ours)  \\
       
        \midrule 
        Train & 0.460 & \textbf{0.131} && 0.460 & \textbf{0.131}  \\
        Val   & 0.475 & \textbf{0.323} && 0.475 & \textbf{0.324}  \\
        Test  & 0.329 & \textbf{0.242} && 0.329 & \textbf{0.242}  \\
    \bottomrule
    \end{tabular}
\label{tab:experiment_perf_elevation}
\end{table}

Overall \glsentrylong{ours} outperforms \glsentrylong{stack}. 
Following the traversability analysis, we present the cell-wise error maps in MAE \figref{fig:experiments-error_maps} (b).
Within both elevation error maps, small artifacts in the form of a unit circle are visible, where a jump in performance decrease can be observed. This artifact can be attributed to the fact that while \glsentrylong{ours} predicts in the vehicle-centric frame, the ground truth is generated using a fixed heading. This artifact is further discussed within \secref{sub:limitations}.

\begin{figure*}[t]
    \centering
    \includegraphics[width=0.85\textwidth]{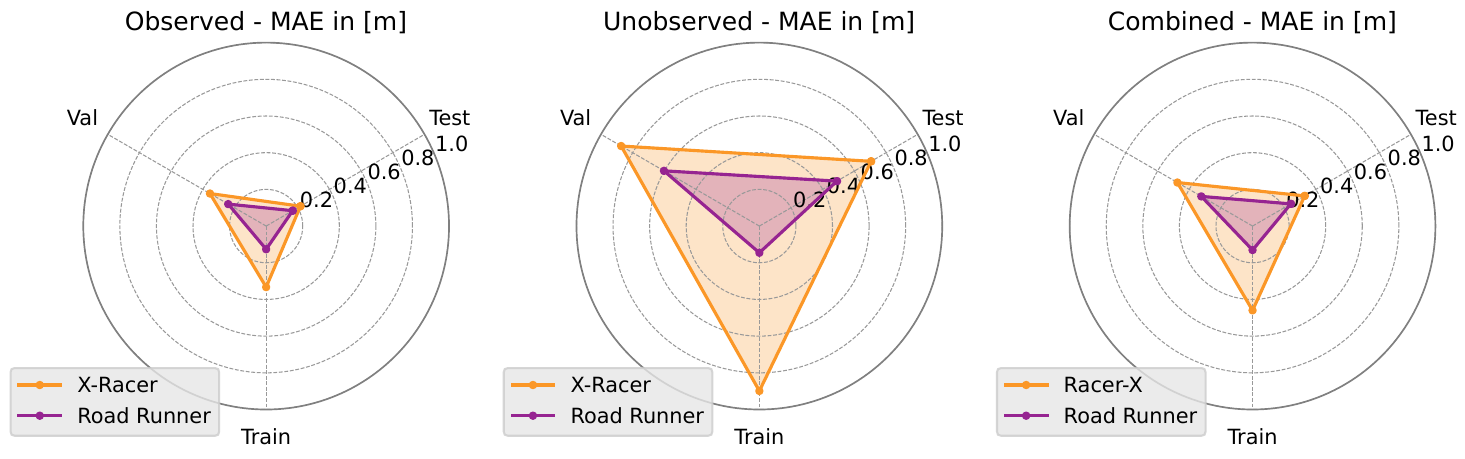}
    \caption{ Comparison of Elevation Mapping Performance across Training, Validation, and Test Datasets. \textit{Observed} indicates that \lidar measurement is available for the predicted cell, while \textit{Unobserved} denotes that no \lidar measurement is available. \textit{Combined} is the evaluation of both  \textit{Unobserved} and \textit{Observed} regions. Note that there are significantly more observed than unobserved regions therefore, \textit{Combined} does not correspond to the arithmetic mean. 
    }
    \label{fig:experiments-radar-plot-elevation}
\end{figure*}

To further understand when \glsentrylong{ours} improves the elevation mapping performance, we split cells into \textit{observed} and \textit{unobserved}. 
We defined the \textit{observability} of a cell based on whether geometric \lidar measurements have been registered for the specific cell or not.  
As shown in \figref{fig:experiments-radar-plot-elevation}, \glsentrylong{ours} strictly outperforms \glsentrylong{stack} across all datasets and for \textit{observed} and \textit{unobserved} areas. 
In areas that are \textit{unobserved}, which tend to be either occluded or far away from the vehicle, our method generally has a higher \gls{mae}. For \textit{observed} regions, the interpolation of \glsentrylong{stack} fails to accurately reconstruct the elevation, while \glsentrylong{ours} shows better performance. However, our method encounters challenges due to domain shift when dealing with the \textit{unobserved} regions in the validation and test environments, given that the \gls{mae} is significantly higher for validation and testing than for the training regions (compare \figref{fig:experiments-radar-plot-elevation} center plot performance difference between \glsentrylong{stack} and RoadRunner). In contrary \glsentrylong{stack} performs consistently across datasets. We hypothesize, that this is due to the relatively small dataset providing evidence that by increasing the data augmentation or the dataset size even better performance can be achieved.  

In summary, we hypothesize that using visual and geometric cues end-to-end can benefit traversability prediction, while accurate prediction of a continuous elevation is significantly harder, specifically when no geometric information is available. 

\subsection{Comparison and Ablation Study}

\begin{table}[t]
\footnotesize
	\centering
	\ra{1.2}
	\setlength{\tabcolsep}{2.4pt}
 	\caption{Comparison and Ablation}
    \begin{tabular}{lcclcccc} \toprule
        & \multicolumn{2}{c}{Elevation} && \multicolumn{4}{c}{Risk MSE}  \\
        \cmidrule{2-3} \cmidrule{5-8}
        & MAE$\downarrow$ & WMAE$\downarrow$ && MSE$\downarrow$ & Recall$\uparrow$ & Precision$\uparrow$ & F1$\uparrow$ \\
       
        \midrule 
        (ours)             & \textbf{0.242} & \textbf{0.241} && \textbf{0.0113} & \textbf{0.302} & \textbf{0.412} & \textbf{0.349} \\
        PointPillar        & 0.645 & 0.643 && 0.0118 & 0.274 & 0.367 & 0.314 \\ 
        LSS                & 0.643 & 0.640 && 0.0118 & 0.281 & 0.370 & 0.319 \\
        \midrule 
        No Weighting       & \textbf{0.241} & 0.241 && 0.0112 & 0.297 & 0.413 & 0.345 \\
        Not Frozen         & 0.247 & 0.246 && 0.0116 & 0.263 & 0.406 & 0.319 \\
        Ele-Classification & 0.247 & 0.246 && 0.0113 & 0.302 & \textbf{0.415} & 0.349 \\
        Common Decoder     & 0.249 & 0.248 && \textbf{0.0108} & \textbf{0.348} & 0.410 & \textbf{0.376} \\
    \bottomrule
    \end{tabular}
\label{tab:experiment_ablation}
\end{table}

To understand the importance of individual modalities we compare our work to PointPillar~\cite{Lang2019} (geometry-only) and \textit{Lift Splat Shoot}~\cite{Philion2022} (vision-only). 
While a comparison to~\cite{Meng2023} would be desirable, the different choice of sensor modalities (stereo-depth compared to ours using \lidar) and the closed source code renders the comparison infeasible. 

In \tabref{tab:experiment_ablation} we report the the MSE and WMSE for the traversability estimation and the MAE and WMAE for the elevation mapping performance. The superior performance of \glsentrylong{ours} aligns with the findings of \cite{Meng2023,Liu2022} that multi-modal sensor fusion results in superior performance. 

In addition, we ablate the individual design choices of \glsentrylong{ours}. 
\textit{No Weighting} indicates removing the balancing for the traversability and elevation loss and using the MSE and MAE, respectively, without weighting. Removing the balancing slightly increases the traversability prediction performance in terms of MSE (0.0113 $\rightarrow$ 0.0112), but leads to an overall worse F1 score (0.349 $\rightarrow$ 0.345). By the increased weighting of high-risk areas, the \glsentrylong{ours} network predicts a more conservative estimate, which is desirable. 
\textit{Not Frozen} allows to update the weights of the EfficientNet-B0 backbone. 
Updating all weights during training leads to better performance on the training dataset, but overall hinders generalization to new environments leading to a worse test score, across all metrics. This validates our design choice of freezing the EfficientNet-B0 backbone. 
\textit{Common-Decoder} changes the dual decoder network architecture to a shared decoder network for the traversability and elevation mapping task. Given that the performance for the elevation mapping task is degraded (0.242 $\rightarrow$ 0.249), we decided on a separate decoder architecture. 
We hypothesize that the features learned to accurately predict traversability and elevation are not complementary.

\subsection{Deployment}
\label{subsec:deployment}
Our \glsentrylong{vehicle} is equipped with a Threadripper 3990x CPU (64 Core 2.9/4.3 GHz), 256 GB RAM, and 4xGeForce RTX 3080 GPUs. We integrate \glsentrylong{ours} using ROS and use a single GPU for inference. 
To achieve low latency and overcome the slow data serialization of the Python ROS wrapper, we write a C++ node using Python bindings to the network implemented in PyTorch. 

This allows \glsentrylong{ours} to run with a latency of 131.85~$\pm$~\SI{2.5}{ms}, compared to the modular \glsentrylong{stack} with an average latency of over \SI{500}{ms}.
The timings are reported over 700 samples with an average of $36,766$ points per sample.

\begin{figure*}[ht]
   \centering
   \includegraphics[width=\textwidth]{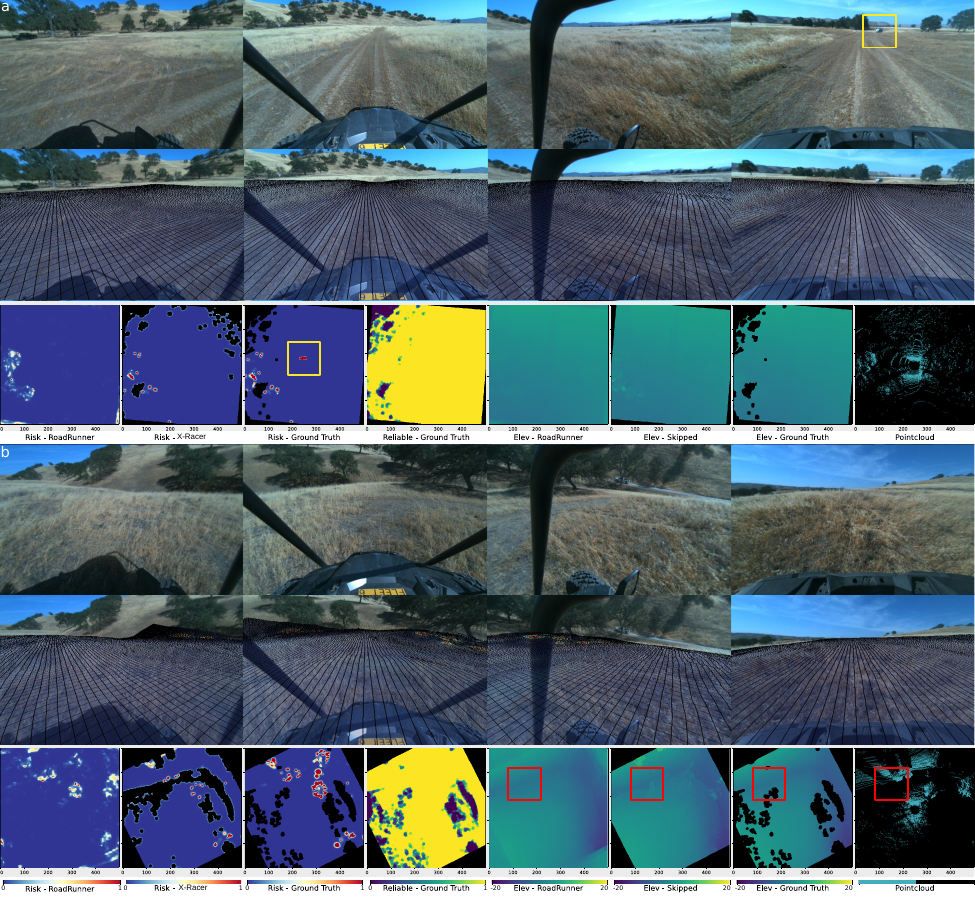}
   \caption{Two Deployment Example Predictions: \textbf{(a)} and \textbf{(b)}. The \textbf{top two rows} of each of the two examples visualize the four onboard camera images (left, front, right, rear). In the \textbf{top row}, the predicted traversability and elevation are projected onto the images. The \textbf{bottom row} shows the respective traversability, reliability, and elevation grid map, as well as the rasterized merged point cloud. The traversability risk maps are shown for \glsentrylong{ours}, \glsentrylong{stack}, and the generated ground truth. \textcolor{blue}{Blue} indicates risk-free traversable and \textcolor{red}{red} untraversable.
   The reliability is shown for the ground truth. The elevation maps are presented for \glsentrylong{ours}, the nearest neighbor interpolated elevation map predicted by \glsentrylong{stack}, and the ground truth elevation. For all maps, black indicates that no prediction is available. \textbf{(a)} Yellow box in \textit{Risk - Ground Truth} indicates a hazard in front of the robot. No hazard is visible in the front-facing camera. The \textit{Risk - Ground Truth} is corrupted by a car driving behind the vehicle visible in the rear camera image. 
   \textbf{(b)} Red box highlights successful elevation prediction, where \glsentrylong{ours} learns to correctly predict the small canyon in front of the vehicle, which is not fully observable from the LiDAR point cloud. Further details and interpretation of each example can be found within \secref{subsec:deployment}.}
    \label{fig:experiments-example_2_outputs}
\end{figure*}

\begin{figure*}[ht]
   \centering
   \includegraphics[width=1.0\textwidth]{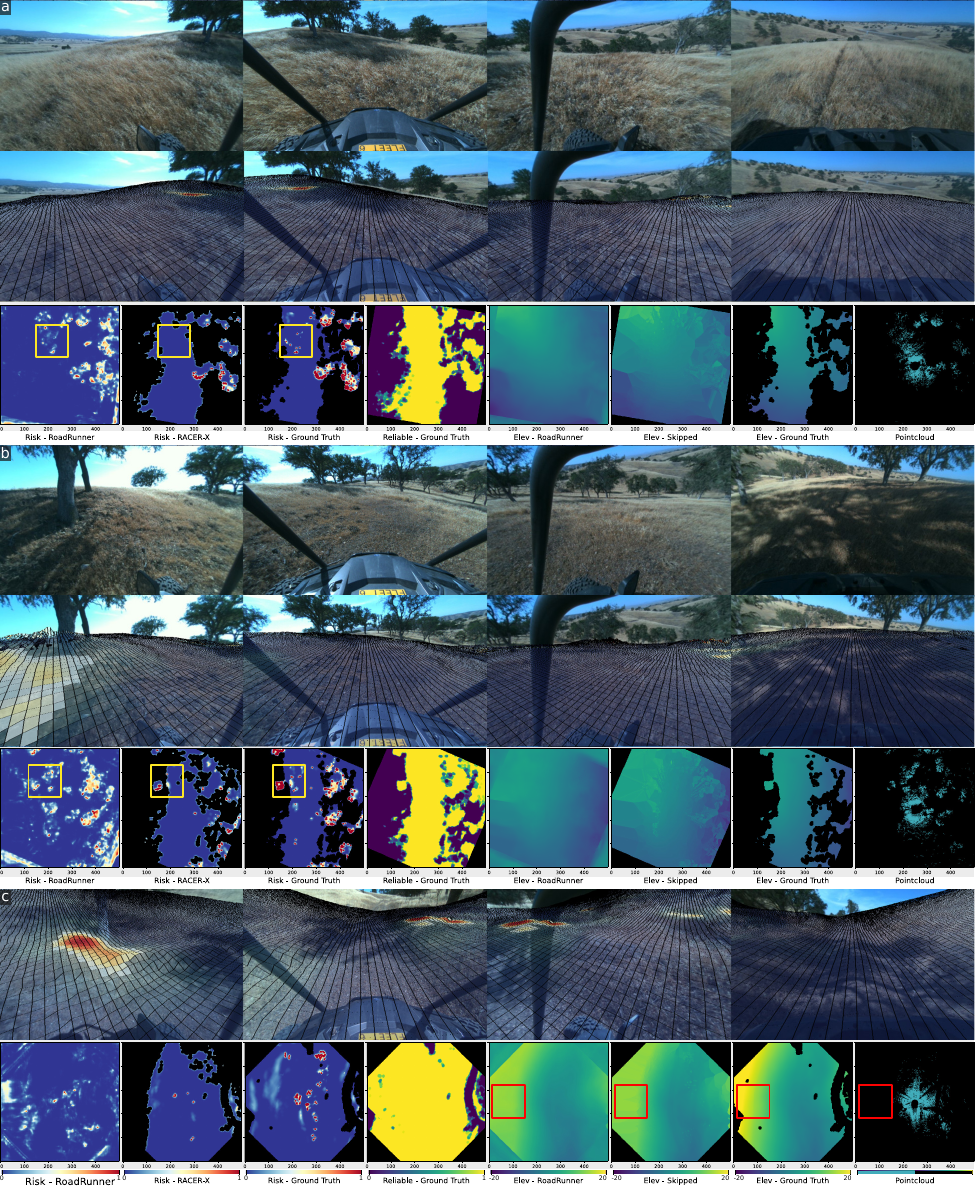}
   \caption{
   \textbf{(a, b)} Yellow box highlights successful predictions at long range where \glsentrylong{stack} fails.
   \textbf{(c)} Red box highlights incorrect elevation prediction, due to missing LiDAR and camera observations.
   Further details and interpretation of each example can be found within \secref{subsec:deployment} and caption of \figref{fig:experiments-example_2_outputs}.}
    \label{fig:experiments-example_3_outputs}
\end{figure*}

\figref{fig:experiments-example_2_outputs} and \figref{fig:experiments-example_3_outputs} provide example outputs of \glsentrylong{ours} illustrating successful predictions as well as the limitations. Each output includes visualizations of the traversability risk predicted by \glsentrylong{ours}, \glsentrylong{stack}, side-by-side to the (hindsight-) generated ground truth risk. Additionally, we present visualizations for our predicted elevation, the feed-forwarded elevation (nearest neighbor interpolated elevation map predicted by \glsentrylong{stack}), and the ground truth elevation map. The currently merged point cloud is displayed, along with an assessment of the reliability of the ground truth. 
In the top row, the predicted traversability and elevation by \glsentrylong{ours} are projected onto the camera images up to a range of roughly \SI{30}{m}.
This comprehensive presentation offers a detailed insight into the capabilities and performance of our \glsentrylong{ours} in terms of traversability risk and elevation predictions.

\figref{fig:experiments-example_3_outputs} (a) illustrates a driving scenario on top of a hill. \glsentrylong{stack} was unable to accumulate sufficient geometric information to correctly identify the trees ahead (highlighted by the yellow box). In contrast, \glsentrylong{ours} provides a more accurate traversability estimation, correctly identifying the trees ahead as high-risk regions.
In \figref{fig:experiments-example_3_outputs} (b), a similar scenario is presented: \glsentrylong{ours} correctly identifies tree trunks as untraversable compared to \glsentrylong{stack}. Specifically, the two trees to the direct left of the vehicle are identified as higher-risk regions (highlighted by the yellow box).
In \figref{fig:experiments-example_3_outputs} (c), the predicted traversability risk map correctly identifies the most prominent risk in the scene -- the two trees in the direct surroundings. On the other hand, the trees behind the vehicle are not identified correctly in the traversability risk map. Concerning elevation, while \glsentrylong{ours} accurately predicts a smooth increase in elevation to the left of the vehicle, it is unable to estimate the correct numerical values and underestimates the slope (highlighted by the red box). This discrepancy becomes apparent when comparing the elevation predicted by \glsentrylong{ours} with the ground truth map. 
We hypothesize that this issue arises due to the limited information available from the images and point cloud, which only covers the bottom part of the valley. As a result, \glsentrylong{ours} can only provide a "guess" of the elevation profile within the unobserved areas, given that no direct image or \lidar measurements are available.

In \figref{fig:experiments-example_2_outputs} (a), we highlighted a limitation of automatic ground truth generation. In this scenario, a second vehicle (visible in the rear-facing camera, highlighted by the yellow box) is following the main vehicle. Consequently, the ground truth risk map is corrupted by the motion of this dynamic obstacle. This can be seen by the small untraversable regions in front of the vehicle within the ground truth traversability risk (highlighted by the yellow box). Despite this, both \glsentrylong{ours} and \glsentrylong{stack} can accurately predict the correct traversability, where the empty road and field ahead are labeled as risk-free traversable.

\afterpage{\clearpage}

In \figref{fig:experiments-example_2_outputs} (b), \glsentrylong{ours} demonstrates a particularly accurate estimation of the scene's elevation. The bottom of the canyon in front of the vehicle, which cannot the sensed by the LiDAR sensors, is nearly perfectly identified.

\subsection{Limitations}
\label{sub:limitations}

The generalization capability of \glsentrylong{ours} to novel environments is constrained by the limited size and diversity of the training dataset. Ongoing efforts focus on scaling up the dataset size. Given the constraints imposed by limited training data, it is not anticipated that \glsentrylong{ours} will seamlessly adapt to entirely new, out-of-distribution environments, and we only showed generalization to data within the same ecoregion.

Moreover, \glsentrylong{ours}'s dependence on self-supervised ground truth generation based on \glsentrylong{stack} imposes limitations on its performance. Although self-supervision is cost-effective and eliminates the need for human annotation, \glsentrylong{ours}'s performance in a sense is upper bounded by \glsentrylong{stack}'s capability to identify risks. The integration of sparse human supervision can help to address the existing failure cases of \glsentrylong{stack}, such as metallic wire fences that remain undetected by LiDARs and are easily misclassified in the image domain. Additionally, dynamic obstacles as seen in \figref{fig:experiments-example_2_outputs} (a) are not correctly handled in the ground truth generation pipeline.
Furthermore, concerning the generation of ground truth, it is important to note that while the \glsentrylong{stack} operates within a yaw-fixed frame, our approach (\glsentrylong{ours}) provides predictions within the robot-centric frame. As a result, aligning the pseudo ground truth to the robot-centric frame leads to artifacts in the form of invalid values depending on the rotation angle between the two frames, as shown in~\figref{fig:experiments-example_2_outputs} and~\figref{fig:experiments-example_3_outputs}. 
Due to the missing supervision during training for those specific regions, the network struggles to learn correct predictions. As part of our future work, we plan to address this implementation detail by expanding the pseudo ground truth maps by a factor of $\sqrt{2}$, to prohibit those artifacts.

Another limitation is the temporal consistency of the predictions of \glsentrylong{ours} as well as the missing memory. For instance, when navigating down into a canyon, \glsentrylong{ours} cannot retain information about obstacles outside the current field of view, as obstacles cannot be captured by cameras nor \lidar measurements (compare \figref{fig:experiments-example_3_outputs} (c)). This deficit could be addressed by adding memory units or (autoregressive) feedback loops to the \glsentrylong{ours} network architecture. 

One aspect we did not address was evaluating prediction performance as a function of vehicle velocity. While RoadRunner is inherently designed for handling high speeds, a more rigorous quantitative evaluation is needed. One e.g., could drive the same trajectory at different speeds and compare the performance of RoadRunner as a function of velocity. 

While \glsentrylong{ours} can provide accurate traversability and elevation information, a better understanding of the uncertainty of the prediction would be beneficial for the planning and fusion of the predictions. For this, one could integrate methods available in the evidential deep learning community to understand the epistemic uncertainty and detect out-of-distribution data specifically.

Lastly, \glsentrylong{ours} does not provide theoretical guarantees, in contrast to \glsentrylong{stack}, where, under specific conditions such as optimal sensing and constrained environments, guarantees can be made based on implemented heuristics.

\section{CONCLUSION \& FUTURE WORK}
\label{sec:Conclusion}
In conclusion, \glsentrylong{ours} represents a significant leap forward in the realm of mapping and traversability estimation for high-speed off-road autonomous robot navigation. We showcase improved traversability prediction and elevation mapping performance, the benefit of sensor fusion, and a rigorous analysis of the performance for a \SI{16.5}{km} off-road dataset.
With a substantial reduction in latency, \glsentrylong{ours} has the potential to enable vehicles to operate at higher speeds autonomously in complex unstructured environments. The implications of this approach extend to critical areas such as search and rescue operations and even hold promise for robot applications in extraterrestrial terrain, where reliable traversability understanding is key.

In future research, we will address existing challenges laid out in the \secref{sub:limitations}, while exploring novel avenues, including attention-based fusion of information, evidential deep learning methods for uncertainty prediction, and a more physically grounded estimation of traversability concerning vehicle dynamics by exploring the integration of the learned dynamics models. We hypothesize novel data augmentation methods allowing for a coherent augmentation of point cloud and image data in combination with synthetic data may have the potential to further substantially improve the performance and generalization. Lastly, we recognize the need for publicly available datasets and benchmarks for off-road terrain understanding, which will allow for comparing various methods across varying off-road environments and ecoregions.


\balance
\bibliographystyle{unsrtnat}
\bibliography{bibliography}

\begin{thebibliography}{91}
\providecommand{\natexlab}[1]{#1}
\providecommand{\url}[1]{\texttt{#1}}
\expandafter\ifx\csname urlstyle\endcsname\relax
  \providecommand{\doi}[1]{doi: #1}\else
  \providecommand{\doi}{doi: \begingroup \urlstyle{rm}\Url}\fi

\bibitem[Maxon(2005)]{maxon2005real}
Martha~Anne Maxon.
\newblock \emph{The real roadrunner}, volume~9.
\newblock University of Oklahoma Press, 2005.

\bibitem[Stanislas et~al.(2021)Stanislas, Nubert, Dugas, Nitsch, S{\"u}nderhauf, Siegwart, Cadena, and Peynot]{stanislas2021airborne}
Leo Stanislas, Julian Nubert, Daniel Dugas, Julia Nitsch, Niko S{\"u}nderhauf, Roland Siegwart, Cesar Cadena, and Thierry Peynot.
\newblock Airborne particle classification in lidar point clouds using deep learning.
\newblock In \emph{Field and Service Robotics: Results of the 12th International Conference}, pages 395--410. Springer, 2021.

\bibitem[Philion and Fidler(2020)]{Philion2022}
Jonah Philion and Sanja Fidler.
\newblock Lift, splat, shoot: Encoding images from arbitrary camera rigs by implicitly unprojecting to 3d.
\newblock In Andrea Vedaldi, Horst Bischof, Thomas Brox, and Jan{-}Michael Frahm, editors, \emph{Computer Vision - {ECCV} 2020 - 16th European Conference, Glasgow, UK, August 23-28, 2020, Proceedings, Part {XIV}}, volume 12359 of \emph{Lecture Notes in Computer Science}, pages 194--210. Springer, 2020.
\newblock \doi{10.1007/978-3-030-58568-6\_12}.

\bibitem[Lang et~al.(2019)Lang, Vora, Caesar, Zhou, Yang, and Beijbom]{Lang2019}
Alex~H. Lang, Sourabh Vora, Holger Caesar, Lubing Zhou, Jiong Yang, and Oscar Beijbom.
\newblock Pointpillars: Fast encoders for object detection from point clouds.
\newblock In \emph{{IEEE} Conference on Computer Vision and Pattern Recognition, {CVPR} 2019, Long Beach, CA, USA, June 16-20, 2019}, pages 12697--12705. Computer Vision Foundation / {IEEE}, 2019.
\newblock \doi{10.1109/CVPR.2019.01298}.

\bibitem[Liu et~al.(2023)Liu, Tang, Amini, Yang, Mao, Rus, and Han]{Liu2022}
Zhijian Liu, Haotian Tang, Alexander Amini, Xingyu Yang, Huizi Mao, Daniela Rus, and Song Han.
\newblock Bevfusion: Multi-task multi-sensor fusion with unified bird's-eye view representation.
\newblock In \emph{IEEE International Conference on Robotics and Automation (ICRA)}, 2023.

\bibitem[Shoemaker and Bornstein(1998)]{Shoemaker1998}
Charles~M Shoemaker and Jonathan~A Bornstein.
\newblock The demo iii ugv program: A testbed for autonomous navigation research.
\newblock In \emph{Proceedings of the 1998 IEEE International Symposium on Intelligent Control (ISIC) held jointly with IEEE International Symposium on Computational Intelligence in Robotics and Automation (CIRA) Intell}, pages 644--651. IEEE, 1998.

\bibitem[Kelly et~al.(2006)Kelly, Stentz, Amidi, Bode, Bradley, Diaz-Calderon, Happold, Herman, Mandelbaum, Pilarski, et~al.]{Kelly2006}
Alonzo Kelly, Anthony Stentz, Omead Amidi, Mike Bode, David Bradley, Antonio Diaz-Calderon, Mike Happold, Herman Herman, Robert Mandelbaum, Tom Pilarski, et~al.
\newblock Toward reliable off road autonomous vehicles operating in challenging environments.
\newblock \emph{The International Journal of Robotics Research}, 25\penalty0 (5-6):\penalty0 449--483, 2006.

\bibitem[Kim et~al.(2006)Kim, Sun, Oh, Rehg, and Bobick]{Kim2006}
Dongshin Kim, Jie Sun, Sang~Min Oh, J.M. Rehg, and A.F. Bobick.
\newblock Traversability classification using unsupervised on-line visual learning for outdoor robot navigation.
\newblock In \emph{Proceedings 2006 IEEE International Conference on Robotics and Automation, 2006. ICRA 2006.}, pages 518--525, 2006.
\newblock \doi{10.1109/ROBOT.2006.1641763}.

\bibitem[Hadsell et~al.(2009)Hadsell, Sermanet, Ben, Erkan, Scoffier, Kavukcuoglu, Muller, and LeCun]{Hadsell2009}
Raia Hadsell, Pierre Sermanet, Jan Ben, Ayse Erkan, Marco Scoffier, Koray Kavukcuoglu, Urs Muller, and Yann LeCun.
\newblock Learning long-range vision for autonomous off-road driving.
\newblock \emph{Journal of Field Robotics}, 26\penalty0 (2):\penalty0 120--144, 2009.
\newblock \doi{https://doi.org/10.1002/rob.20276}.

\bibitem[Guzzi et~al.(2020)Guzzi, Chavez-Garcia, Nava, Gambardella, and Giusti]{Guzzi20}
Jérôme Guzzi, R.~Omar Chavez-Garcia, Mirko Nava, Luca~Maria Gambardella, and Alessandro Giusti.
\newblock Path planning with local motion estimations.
\newblock \emph{IEEE Robotics and Automation Letters}, 5\penalty0 (2):\penalty0 2586--2593, 2020.
\newblock \doi{10.1109/LRA.2020.2972849}.

\bibitem[Frey et~al.(2022)Frey, Hoeller, Khattak, and Hutter]{Frey2022}
Jonas Frey, David Hoeller, Shehryar Khattak, and Marco Hutter.
\newblock Locomotion policy guided traversability learning using volumetric representations of complex environments.
\newblock In \emph{2022 IEEE/RSJ International Conference on Intelligent Robots and Systems (IROS)}. IEEE, 2022.

\bibitem[Frey et~al.(2023)Frey, Mattamala, Chebrolu, Cadena, Fallon, and Hutter]{Frey2023}
Jonas Frey, Matias Mattamala, Nived Chebrolu, Cesar Cadena, Maurice Fallon, and Marco Hutter.
\newblock {Fast Traversability Estimation for Wild Visual Navigation}.
\newblock In \emph{Proceedings of Robotics: Science and Systems}, Daegu, Republic of Korea, July 2023.
\newblock \doi{10.15607/RSS.2023.XIX.054}.

\bibitem[Fan et~al.(2021)Fan, Otsu, Kubo, Dixit, Burdick, and Agha{-}Mohammadi]{Fan2021}
David~D. Fan, Kyohei Otsu, Yuki Kubo, Anushri Dixit, Joel Burdick, and Ali{-}Akbar Agha{-}Mohammadi.
\newblock {STEP:} stochastic traversability evaluation and planning for safe off-road navigation.
\newblock \emph{CoRR}, abs/2103.02828, 2021.

\bibitem[Majumdar and Pavone(2017)]{Majumdar2017}
Anirudha Majumdar and Marco Pavone.
\newblock How should a robot assess risk? towards an axiomatic theory of risk in robotics.
\newblock In Nancy~M. Amato, Greg Hager, Shawna~L. Thomas, and Miguel Torres{-}Torriti, editors, \emph{Robotics Research, The 18th International Symposium, {ISRR} 2017, Puerto Varas, Chile, December 11-14, 2017}, volume~10 of \emph{Springer Proceedings in Advanced Robotics}, pages 75--84. Springer, 2017.
\newblock \doi{10.1007/978-3-030-28619-4\_10}.

\bibitem[Pomerleau(1993)]{Pomerleau1993}
Dean~A Pomerleau.
\newblock Knowledge-based training of artificial neural networks for autonomous robot driving.
\newblock In \emph{Robot learning}, pages 19--43. Springer, 1993.

\bibitem[Manduchi et~al.(2005)Manduchi, Castano, Talukder, and Matthies]{Manduchi2005}
Roberto Manduchi, Andres Castano, Ashit Talukder, and Larry Matthies.
\newblock Obstacle detection and terrain classification for autonomous off-road navigation.
\newblock \emph{Autonomous robots}, 18:\penalty0 81--102, 2005.

\bibitem[Jackel et~al.(2006)Jackel, Krotkov, Perschbacher, Pippine, and Sullivan]{Jackel2006}
L.~D. Jackel, Eric Krotkov, Michael Perschbacher, Jim Pippine, and Chad Sullivan.
\newblock The darpa lagr program: Goals, challenges, methodology, and phase i results.
\newblock \emph{Journal of Field Robotics}, 23\penalty0 (11-12):\penalty0 945--973, 2006.
\newblock \doi{https://doi.org/10.1002/rob.20161}.

\bibitem[Muller et~al.(2005)Muller, Ben, Cosatto, Flepp, and Cun]{Muller2005}
Urs Muller, Jan Ben, Eric Cosatto, Beat Flepp, and Yann Cun.
\newblock Off-road obstacle avoidance through end-to-end learning.
\newblock In Y.~Weiss, B.~Sch\"{o}lkopf, and J.~Platt, editors, \emph{Advances in Neural Information Processing Systems}, volume~18. MIT Press, 2005.

\bibitem[Hadsell et~al.(2007)Hadsell, Sermanet, Ben, Erkan, Han, Flepp, Muller, and LeCun]{Hadsell2007}
Raia Hadsell, Pierre Sermanet, Jan Ben, A~Erkan, Jeff Han, Beat Flepp, Urs Muller, and Yann LeCun.
\newblock Online learning for offroad robots: Using spatial label propagation to learn long-range traversability.
\newblock In \emph{Proc. of Robotics: Science and Systems (RSS)}, volume~11, page~32. Citeseer, 2007.

\bibitem[Konolige et~al.(2009)Konolige, Agrawal, Blas, Bolles, Gerkey, Sola, and Sundaresan]{Konolige2009}
Kurt Konolige, Motilal Agrawal, Morten~Rufus Blas, Robert~C Bolles, Brian Gerkey, Joan Sola, and Aravind Sundaresan.
\newblock Mapping, navigation, and learning for off-road traversal.
\newblock \emph{Journal of Field Robotics}, 26\penalty0 (1):\penalty0 88--113, 2009.

\bibitem[Behringer et~al.(2004)Behringer, Sundareswaran, Gregory, Elsley, Addison, Guthmiller, Daily, and Bevly]{Behringer2004}
Reinhold Behringer, Sundar Sundareswaran, Brian Gregory, Richard Elsley, Bob Addison, Wayne Guthmiller, Robert Daily, and David Bevly.
\newblock The darpa grand challenge-development of an autonomous vehicle.
\newblock In \emph{IEEE Intelligent Vehicles Symposium, 2004}, pages 226--231. IEEE, 2004.

\bibitem[Thrun et~al.(2006)Thrun, Montemerlo, and Aron]{Thrun2006}
Sebastian Thrun, Michael Montemerlo, and Andrei Aron.
\newblock Probabilistic terrain analysis for high-speed desert driving.
\newblock In \emph{Robotics: Science and Systems}, pages 16--19, 2006.

\bibitem[Urmson et~al.(2006)Urmson, Ragusa, Ray, Anhalt, Bartz, Galatali, Gutierrez, Johnston, Harbaugh, “Yu”~Kato, et~al.]{Urmson2006}
Chris Urmson, Charlie Ragusa, David Ray, Joshua Anhalt, Daniel Bartz, Tugrul Galatali, Alexander Gutierrez, Josh Johnston, Sam Harbaugh, Hiroki “Yu”~Kato, et~al.
\newblock A robust approach to high-speed navigation for unrehearsed desert terrain.
\newblock \emph{Journal of Field Robotics}, 23\penalty0 (8):\penalty0 467--508, 2006.

\bibitem[Tan and Le(2019)]{Tan19}
Mingxing Tan and Quoc~V. Le.
\newblock Efficientnet: Rethinking model scaling for convolutional neural networks.
\newblock In Kamalika Chaudhuri and Ruslan Salakhutdinov, editors, \emph{Proceedings of the 36th International Conference on Machine Learning, {ICML} 2019, 9-15 June 2019, Long Beach, California, {USA}}, volume~97 of \emph{Proceedings of Machine Learning Research}, pages 6105--6114. {PMLR}, 2019.

\bibitem[Viswanath et~al.(2021)Viswanath, Singh, Jiang, Sujit, and Saripalli]{Viswanath2021}
Kasi Viswanath, Kartikeya Singh, Peng Jiang, PB~Sujit, and Srikanth Saripalli.
\newblock Offseg: A semantic segmentation framework for off-road driving.
\newblock In \emph{2021 IEEE 17th International Conference on Automation Science and Engineering (CASE)}, pages 354--359. IEEE, 2021.

\bibitem[Maturana et~al.(2017)Maturana, Chou, Uenoyama, and Scherer]{Maturana2017}
Daniel Maturana, Po-Wei Chou, Masashi Uenoyama, and Sebastian Scherer.
\newblock Real-time semantic mapping for autonomous off-road navigation.
\newblock In \emph{Proceedings of 11th International Conference on Field and Service Robotics (FSR '17)}, pages 335 -- 350, September 2017.

\bibitem[Roth et~al.(2023)Roth, Nubert, Yang, Mittal, and Hutter]{Roth2023}
Pascal Roth, Julian Nubert, Fan Yang, Mayank Mittal, and Marco Hutter.
\newblock Viplanner: Visual semantic imperative learning for local navigation.
\newblock In \emph{2023 IEEE International Conference on Robotics and Automation (ICRA)}, 2023.

\bibitem[Shaban et~al.(2022)Shaban, Meng, Lee, Boots, and Fox]{Shaban2022}
Amirreza Shaban, Xiangyun Meng, JoonHo Lee, Byron Boots, and Dieter Fox.
\newblock Semantic terrain classification for off-road autonomous driving.
\newblock In \emph{Conference on Robot Learning}, pages 619--629. PMLR, 2022.

\bibitem[Erni et~al.(2023)Erni, Frey, Miki, Mattamala~Aravena, and Hutter]{Gian2023}
Gian Erni, Jonas Frey, Takahiro Miki, Matias~Eduardo Mattamala~Aravena, and Marco Hutter.
\newblock Mem: Multi-modal elevation mapping for robotics and learning.
\newblock In \emph{36th IEEE/RSJ International Conference on Intelligent Robots and Systems (IROS 2023); Conference Location: Detroit, MI, USA; Conference Date: October 1-5, 2023; Conference lecture held on October 4, 2023.}, 2023.

\bibitem[Wigness et~al.(2019)Wigness, Eum, Rogers, Han, and Kwon]{Wigness2019rugd}
Maggie Wigness, Sungmin Eum, John~G Rogers, David Han, and Heesung Kwon.
\newblock A rugd dataset for autonomous navigation and visual perception in unstructured outdoor environments.
\newblock In \emph{2019 IEEE/RSJ International Conference on Intelligent Robots and Systems (IROS)}, pages 5000--5007. IEEE, 2019.

\bibitem[Jiang et~al.(2021)Jiang, Osteen, Wigness, and Saripalli]{Jiang2021rellis}
Peng Jiang, Philip Osteen, Maggie Wigness, and Srikanth Saripalli.
\newblock Rellis-3d dataset: Data, benchmarks and analysis.
\newblock In \emph{2021 IEEE international conference on robotics and automation (ICRA)}, pages 1110--1116. IEEE, 2021.

\bibitem[Valada et~al.(2016)Valada, Oliveira, Brox, and Burgard]{Valada16freiburg}
Abhinav Valada, Gabriel Oliveira, Thomas Brox, and Wolfram Burgard.
\newblock Deep multispectral semantic scene understanding of forested environments using multimodal fusion.
\newblock In \emph{International Symposium on Experimental Robotics (ISER)}, 2016.

\bibitem[Bradley et~al.(2015)Bradley, Chang, Silver, Powers, Herman, Rander, and Stentz]{Bradley2015}
David~M. Bradley, Jonathan~K. Chang, David Silver, Matthew Powers, Herman Herman, Peter Rander, and Anthony Stentz.
\newblock Scene understanding for a high-mobility walking robot.
\newblock In \emph{2015 IEEE/RSJ International Conference on Intelligent Robots and Systems (IROS)}, pages 1144--1151, 2015.
\newblock \doi{10.1109/IROS.2015.7353514}.

\bibitem[Schilling et~al.(2017)Schilling, Chen, Folkesson, and Jensfelt]{Schilling2017}
Fabian Schilling, Xi~Chen, John Folkesson, and Patric Jensfelt.
\newblock Geometric and visual terrain classification for autonomous mobile navigation.
\newblock In \emph{2017 IEEE/RSJ International Conference on Intelligent Robots and Systems (IROS)}, pages 2678--2684, 2017.
\newblock \doi{10.1109/IROS.2017.8206092}.

\bibitem[Ono et~al.(2015)Ono, Fuchs, Steffy, Maimone, and Yen]{Ono2015}
Masahiro Ono, Thoams~J Fuchs, Amanda Steffy, Mark Maimone, and Jeng Yen.
\newblock Risk-aware planetary rover operation: Autonomous terrain classification and path planning.
\newblock In \emph{2015 IEEE aerospace conference}, pages 1--10. IEEE, 2015.

\bibitem[Rothrock et~al.(2016)Rothrock, Kennedy, Cunningham, Papon, Heverly, and Ono]{Rothrock2016}
Brandon Rothrock, Ryan Kennedy, Chris Cunningham, Jeremie Papon, Matthew Heverly, and Masahiro Ono.
\newblock \emph{Spoc: Deep learning-based terrain classification for mars rover missions}, page 5539.
\newblock Aerospace Research Central, 2016.
\newblock \doi{10.2514/6.2016-5539}.
\newblock URL \url{https://arc.aiaa.org/doi/abs/10.2514/6.2016-5539}.

\bibitem[Swan et~al.(2021)Swan, Atha, Leopold, Gildner, Oij, Chiu, and Ono]{Swan2021}
R~Michael Swan, Deegan Atha, Henry~A Leopold, Matthew Gildner, Stephanie Oij, Cindy Chiu, and Masahiro Ono.
\newblock Ai4mars: A dataset for terrain-aware autonomous driving on mars.
\newblock In \emph{Proceedings of the IEEE/CVF Conference on Computer Vision and Pattern Recognition}, pages 1982--1991, 2021.

\bibitem[Zhang et~al.(2022)Zhang, Lin, Fan, Wang, and Liu]{Zhang2022}
Jiahang Zhang, Lilang Lin, Zejia Fan, Wenjing Wang, and Jiaying Liu.
\newblock S$^5$mars: Self-supervised and semi-supervised learning for mars segmentation.
\newblock \emph{arXiv preprint arXiv:2207.01200}, 2022.

\bibitem[Goh et~al.(2022)Goh, Chen, and Wilson]{Goh2022}
Edwin Goh, Jingdao Chen, and Brian Wilson.
\newblock Mars terrain segmentation with less labels.
\newblock In \emph{2022 IEEE Aerospace Conference (AERO)}, pages 1--10. IEEE, 2022.

\bibitem[Endo et~al.(2023)Endo, Taniai, Yonetani, and Ishigami]{Endo2023}
Masafumi Endo, Tatsunori Taniai, Ryo Yonetani, and Genya Ishigami.
\newblock Risk-aware path planning via probabilistic fusion of traversability prediction for planetary rovers on heterogeneous terrains.
\newblock \emph{arXiv preprint arXiv:2303.01169}, 2023.

\bibitem[Meng et~al.(2023)Meng, Hatch, Lambert, Li, Wagener, Schmittle, Lee, Yuan, Chen, Deng, Okopal, Fox, Boots, and Shaban]{Meng2023}
Xiangyun Meng, Nathan Hatch, Alexander Lambert, Anqi Li, Nolan Wagener, Matthew Schmittle, JoonHo Lee, Wentao Yuan, Zoey Chen, Samuel Deng, Greg Okopal, Dieter Fox, Byron Boots, and Amirreza Shaban.
\newblock Terrainnet: Visual modeling of complex terrain for high-speed, off-road navigation.
\newblock In \emph{Proceedings of Robotics: Science and System XIX}. Robotics Science \& Systems Foundation, 2023.
\newblock Robotics: Science and Systems (RSS 2023); Conference Location: Daegu, South Korea; Conference Date: July 10-14, 2023.

\bibitem[Triebel et~al.(2006)Triebel, Pfaff, and Burgard]{Triebel2006}
Rudolph Triebel, Patrick Pfaff, and Wolfram Burgard.
\newblock Multi-level surface maps for outdoor terrain mapping and loop closing.
\newblock In \emph{2006 IEEE/RSJ International Conference on Intelligent Robots and Systems}, pages 2276--2282, 2006.
\newblock \doi{10.1109/IROS.2006.282632}.

\bibitem[Brooks and Iagnemma(2012)]{Brooks2012}
Christopher~A Brooks and Karl Iagnemma.
\newblock Self-supervised terrain classification for planetary surface exploration rovers.
\newblock \emph{Journal of Field Robotics}, 29\penalty0 (3):\penalty0 445--468, 2012.

\bibitem[Otsu et~al.(2016)Otsu, Ono, Fuchs, Baldwin, and Kubota]{Otsu2016}
Kyohei Otsu, Masahiro Ono, Thomas~J. Fuchs, Ian Baldwin, and Takashi Kubota.
\newblock Autonomous terrain classification with co- and self-training approach.
\newblock \emph{IEEE Robotics and Automation Letters}, 1\penalty0 (2):\penalty0 814--819, 2016.
\newblock \doi{10.1109/LRA.2016.2525040}.

\bibitem[Castro et~al.(2023)Castro, Triest, Wang, Gregory, Sanchez, Rogers, and Scherer]{Castro2023}
Mateo~Guaman Castro, Samuel Triest, Wenshan Wang, Jason~M Gregory, Felix Sanchez, John~G Rogers, and Sebastian Scherer.
\newblock How does it feel? self-supervised costmap learning for off-road vehicle traversability.
\newblock In \emph{2023 IEEE International Conference on Robotics and Automation (ICRA)}, pages 931--938. IEEE, 2023.

\bibitem[Seo et~al.(2023)Seo, Sim, and Shim]{Seo2023learning}
Junwon Seo, Sungdae Sim, and Inwook Shim.
\newblock Learning off-road terrain traversability with self-supervisions only.
\newblock \emph{IEEE Robotics and Automation Letters}, 2023.

\bibitem[Higa et~al.(2019)Higa, Iwashita, Otsu, Ono, Lamarre, Didier, and Hoffmann]{Higa2019}
Shoya Higa, Yumi Iwashita, Kyohei Otsu, Masahiro Ono, Olivier Lamarre, Annie Didier, and Mark Hoffmann.
\newblock Vision-based estimation of driving energy for planetary rovers using deep learning and terramechanics.
\newblock \emph{IEEE Robotics and Automation Letters}, 4\penalty0 (4):\penalty0 3876--3883, 2019.

\bibitem[Z{\"u}rn et~al.(2021)Z{\"u}rn, Burgard, and Valada]{Zurn2021}
Jannik Z{\"u}rn, Wolfram Burgard, and Abhinav Valada.
\newblock Self-supervised visual terrain classification from unsupervised acoustic feature learning.
\newblock \emph{IEEE Transactions on Robotics}, 37\penalty0 (2):\penalty0 466--481, 2021.

\bibitem[Sathyamoorthy et~al.(2022)Sathyamoorthy, Weerakoon, Guan, Liang, and Manocha]{Sathyamoorthy2022}
Adarsh~Jagan Sathyamoorthy, Kasun Weerakoon, Tianrui Guan, Jing Liang, and Dinesh Manocha.
\newblock Terrapn: Unstructured terrain navigation using online self-supervised learning.
\newblock In \emph{{IEEE/RSJ} International Conference on Intelligent Robots and Systems, {IROS} 2022, Kyoto, Japan, October 23-27, 2022}, pages 7197--7204. {IEEE}, 2022.
\newblock \doi{10.1109/IROS47612.2022.9981942}.

\bibitem[Richter and Roy(2017)]{Richter17}
Charles Richter and Nicholas Roy.
\newblock Safe visual navigation via deep learning and novelty detection.
\newblock In \emph{Robotics: Science and Systems}, Cambridge, Massachusetts, July 2017.
\newblock \doi{10.15607/RSS.2017.XIII.064}.

\bibitem[Seo et~al.(2022)Seo, Kim, Kwak, Min, and Shim]{Seo2022}
Junwon Seo, Taekyung Kim, Ki~Ho Kwak, Jihong Min, and Inwook Shim.
\newblock Scate: A scalable framework for self- supervised traversability estimation in unstructured environments.
\newblock \emph{IEEE Robotics and Automation Letters}, 8:\penalty0 888--895, 2022.

\bibitem[Ahtiainen et~al.(2017)Ahtiainen, Stoyanov, and Saarinen]{Ahtiainen2017}
Juhana Ahtiainen, Todor Stoyanov, and Jari Saarinen.
\newblock Normal distributions transform traversability maps: lidar-only approach for traversability mapping in outdoor environments.
\newblock \emph{Journal of Field Robotics}, 34\penalty0 (3):\penalty0 600--621, 2017.

\bibitem[Gasparino et~al.(2022)Gasparino, Sivakumar, Liu, Velasquez, Higuti, Rogers, Tran, and Chowdhary]{Gasparino2022}
Mateus~V. Gasparino, Arun~N. Sivakumar, Yixiao Liu, Andres E.~B. Velasquez, Vitor A.~H. Higuti, John Rogers, Huy Tran, and Girish Chowdhary.
\newblock {WayFAST}: Navigation with predictive traversability in the field.
\newblock \emph{{IEEE} Robotics and Automation Letters}, 7\penalty0 (4):\penalty0 10651--10658, oct 2022.
\newblock \doi{10.1109/lra.2022.3193464}.

\bibitem[Cai et~al.(2022)Cai, Everett, Sharma, Osteen, and How]{Cai2022}
Xiaoyi Cai, Michael Everett, Lakshay Sharma, Philip~R Osteen, and Jonathan~P How.
\newblock Probabilistic traversability model for risk-aware motion planning in off-road environments.
\newblock \emph{arXiv preprint arXiv:2210.00153}, 2022.

\bibitem[Xue et~al.(2023{\natexlab{a}})Xue, Hu, Xie, Fu, Xiao, Nie, and Dai]{Xue2023contrastive}
Hanzhang Xue, Xiaochang Hu, Rui Xie, Hao Fu, Liang Xiao, Yiming Nie, and Bin Dai.
\newblock Contrastive label disambiguation for self-supervised terrain traversability learning in off-road environments.
\newblock \emph{arXiv preprint arXiv:2307.02871}, 2023{\natexlab{a}}.

\bibitem[Cai et~al.(2023)Cai, Ancha, Sharma, Osteen, Bucher, Phillips, Wang, Everett, Roy, and How]{Cai2023evora}
Xiaoyi Cai, Siddharth Ancha, Lakshay Sharma, Philip~R. Osteen, Bernadette Bucher, Stephen Phillips, Jiuguang Wang, Michael Everett, Nicholas Roy, and Jonathan~P. How.
\newblock Evora: Deep evidential traversability learning for risk-aware off-road autonomy.
\newblock \emph{arXiv preprint arXiv:2311.06234}, 2023.

\bibitem[Jung et~al.(2023)Jung, Lee, Meng, Boots, and Lambert]{Jung2023}
Sanghun Jung, JoonHo Lee, Xiangyun Meng, Byron Boots, and Alexander Lambert.
\newblock V-strong: Visual self-supervised traversability learning for off-road navigation.
\newblock \emph{arXiv preprint arXiv:2312.16016}, 2023.

\bibitem[Schmid et~al.(2022)Schmid, Atha, Sch{\"o}ller, Dey, Fakoorian, Otsu, Ridge, Bjelonic, Wellhausen, Hutter, et~al.]{Schmid2022}
Robin Schmid, Deegan Atha, Frederik Sch{\"o}ller, Sharmita Dey, Seyed Fakoorian, Kyohei Otsu, Barry Ridge, Marko Bjelonic, Lorenz Wellhausen, Marco Hutter, et~al.
\newblock Self-supervised traversability prediction by learning to reconstruct safe terrain.
\newblock In \emph{2022 IEEE/RSJ International Conference on Intelligent Robots and Systems (IROS)}, pages 12419--12425. IEEE, 2022.

\bibitem[Wellhausen et~al.(2020)Wellhausen, Ranftl, and Hutter]{Wellhausen2020}
Lorenz Wellhausen, René Ranftl, and Marco Hutter.
\newblock {Safe Robot Navigation Via Multi-Modal Anomaly Detection}.
\newblock \emph{IEEE Robotics and Automation Letters}, 2020.
\newblock \doi{10.1109/LRA.2020.2967706}.

\bibitem[Chen et~al.(2023)Chen, Ho, Maulimov, Wang, and Scherer]{Chen2023}
Eric Chen, Cherie Ho, Mukhtar Maulimov, Chen Wang, and Sebastian Scherer.
\newblock Learning-on-the-drive: Self-supervised adaptation of visual offroad traversability models.
\newblock \emph{arXiv preprint arXiv:2306.15226}, 2023.

\bibitem[Wellhausen and Hutter(2023)]{Wellhausen2023}
Lorenz Wellhausen and Marco Hutter.
\newblock Artplanner: Robust legged robot navigation in the field.
\newblock In \emph{Field Robotics}, 2023.

\bibitem[Cao et~al.(2022)Cao, Zhu, Yang, Xia, Choset, Oh, and Zhang]{Cao2022}
Chao Cao, Hongbiao Zhu, Fan Yang, Yukun Xia, Howie Choset, Jean Oh, and Ji~Zhang.
\newblock Autonomous exploration development environment and the planning algorithms.
\newblock In \emph{2022 International Conference on Robotics and Automation (ICRA)}, page 8921–8928. IEEE, 2022.
\newblock \doi{10.1109/ICRA46639.2022.9812330}.

\bibitem[Xue et~al.(2023{\natexlab{b}})Xue, Fu, Xiao, Fan, Zhao, and Dai]{Xue2023}
Hanzhang Xue, Hao Fu, Liang Xiao, Yiming Fan, Dawei Zhao, and Bin Dai.
\newblock Traversability analysis for autonomous driving in complex environment: A lidar-based terrain modeling approach.
\newblock \emph{Journal of Field Robotics}, 2023{\natexlab{b}}.

\bibitem[Hudson et~al.(2022)Hudson, Talbot, Cox, Williams, Hines, Pitt, Wood, Frousheger, Surdo, Molnar, Steindl, Wildie, Sa, Kottege, Stepanas, Hernandez, Catt, Docherty, Tidd, Tam, Murrell, Bessell, Hanson, Tychsen-Smith, Suzuki, Overs, Kendoul, Wagner, Palmer, Milani, O'Brien, Jiang, Chen, and Arkin]{Hudson2022}
Nicolas Hudson, Fletcher Talbot, Mark Cox, Jason Williams, Thomas Hines, Alex Pitt, Brett Wood, Dennis Frousheger, Katrina~Lo Surdo, Thomas Molnar, Ryan Steindl, Matt Wildie, Inkyu Sa, Navinda Kottege, Kazys Stepanas, Emili Hernandez, Gavin Catt, William Docherty, Brendan Tidd, Benjamin Tam, Simon Murrell, Mitchell Bessell, Lauren Hanson, Lachlan Tychsen-Smith, Hajime Suzuki, Leslie Overs, Farid Kendoul, Glenn Wagner, Duncan Palmer, Peter Milani, Matthew O'Brien, Shu Jiang, Shengkang Chen, and Ronald Arkin.
\newblock Heterogeneous ground and air platforms, homogeneous sensing: Team {CSIRO} data61's approach to the {DARPA} subterranean challenge.
\newblock \emph{Field Robotics}, 2\penalty0 (1):\penalty0 595--636, mar 2022.
\newblock \doi{10.55417/fr.2022021}.

\bibitem[Bouman et~al.(2020)Bouman, Ginting, Alatur, Palieri, Fan, Touma, Pailevanian, Kim, Otsu, Burdick, et~al.]{Bouman2020}
Amanda Bouman, Muhammad~Fadhil Ginting, Nikhilesh Alatur, Matteo Palieri, David~D Fan, Thomas Touma, Torkom Pailevanian, Sung-Kyun Kim, Kyohei Otsu, Joel Burdick, et~al.
\newblock Autonomous spot: Long-range autonomous exploration of extreme environments with legged locomotion.
\newblock In \emph{2020 IEEE/RSJ International Conference on Intelligent Robots and Systems (IROS)}, pages 2518--2525. IEEE, 2020.

\bibitem[Kahn et~al.(2021)Kahn, Abbeel, and Levine]{Kahn2021}
Gregory Kahn, Pieter Abbeel, and Sergey Levine.
\newblock {BADGR: An Autonomous Self-Supervised Learning-Based Navigation System}.
\newblock \emph{IEEE Robotics and Automation Letters}, 6\penalty0 (2):\penalty0 1312--1319, 2021.
\newblock \doi{10.1109/LRA.2021.3057023}.

\bibitem[Kim et~al.(2022)Kim, Kim, and Hwangbo]{Kim2022learning}
Yunho Kim, Chanyoung Kim, and Jemin Hwangbo.
\newblock Learning forward dynamics model and informed trajectory sampler for safe quadruped navigation.
\newblock In Kris Hauser, Dylan~A. Shell, and Shoudong Huang, editors, \emph{Robotics: Science and Systems XVIII, New York City, NY, USA, June 27 - July 1, 2022}, 2022.
\newblock \doi{10.15607/RSS.2022.XVIII.069}.

\bibitem[Xiao et~al.(2021)Xiao, Biswas, and Stone]{Xiao2021}
Xuesu Xiao, Joydeep Biswas, and Peter Stone.
\newblock Learning inverse kinodynamics for accurate high-speed off-road navigation on unstructured terrain.
\newblock \emph{IEEE Robotics and Automation Letters}, 6\penalty0 (3):\penalty0 6054--6060, 2021.

\bibitem[Karnan et~al.(2022)Karnan, Sikand, Atreya, Rabiee, Xiao, Warnell, Stone, and Biswas]{Karnan2022}
Haresh Karnan, Kavan~Singh Sikand, Pranav Atreya, Sadegh Rabiee, Xuesu Xiao, Garrett Warnell, Peter Stone, and Joydeep Biswas.
\newblock Vi-ikd: High-speed accurate off-road navigation using learned visual-inertial inverse kinodynamics.
\newblock In \emph{2022 IEEE/RSJ International Conference on Intelligent Robots and Systems (IROS)}, pages 3294--3301. IEEE, 2022.

\bibitem[Wulfmeier et~al.(2015)Wulfmeier, Ondruska, and Posner]{Wulfmeier2015maximum}
Markus Wulfmeier, Peter Ondruska, and Ingmar Posner.
\newblock Maximum entropy deep inverse reinforcement learning.
\newblock \emph{arXiv preprint arXiv:1507.04888}, 2015.

\bibitem[Triest et~al.(2023)Triest, Castro, Maheshwari, Sivaprakasam, Wang, and Scherer]{Triest2023}
Samuel Triest, Mateo~Guaman Castro, Parv Maheshwari, Matthew Sivaprakasam, Wenshan Wang, and Sebastian~A. Scherer.
\newblock Learning risk-aware costmaps via inverse reinforcement learning for off-road navigation.
\newblock \emph{2023 IEEE International Conference on Robotics and Automation (ICRA)}, pages 924--930, 2023.

\bibitem[Fan et~al.(2022)Fan, Agha-mohammadi, and Theodorou]{Fan2022}
David~D. Fan, Ali-akbar Agha-mohammadi, and Evangelos~A. Theodorou.
\newblock Learning risk-aware costmaps for traversability in challenging environments.
\newblock \emph{IEEE Robotics and Automation Letters}, 7\penalty0 (1):\penalty0 279--286, 2022.
\newblock \doi{10.1109/LRA.2021.3125047}.

\bibitem[Dixit et~al.(2023)Dixit, Fan, Otsu, Dey, Agha-Mohammadi, and Burdick]{Dixit2023step}
Anushri Dixit, David~D. Fan, Kyohei Otsu, Sharmita Dey, Ali-Akbar Agha-Mohammadi, and Joel~W. Burdick.
\newblock Step: Stochastic traversability evaluation and planning for risk-aware off-road navigation; results from the darpa subterranean challenge, 2023.

\bibitem[Roddick et~al.(2019)Roddick, Kendall, and Cipolla]{Roddick2019}
Thomas Roddick, Alex Kendall, and Roberto Cipolla.
\newblock Orthographic feature transform for monocular 3d object detection.
\newblock In \emph{30th British Machine Vision Conference 2019, {BMVC} 2019, Cardiff, UK, September 9-12, 2019}, page 285. {BMVA} Press, 2019.

\bibitem[Ng et~al.(2020)Ng, Radia, Chen, Wang, Gog, and Gonzalez]{Ng2022}
Mong~H. Ng, Kaahan Radia, Jianfei Chen, Dequan Wang, Ionel Gog, and Joseph~E. Gonzalez.
\newblock Bev-seg: Bird's eye view semantic segmentation using geometry and semantic point cloud.
\newblock \emph{CVPR 2020 Workshop Scalability in Autonomous Driving by Waymo}, abs/2006.11436, 2020.

\bibitem[Harley et~al.(2023)Harley, Fang, Li, Ambrus, and Fragkiadaki]{Harley2022}
Adam~W. Harley, Zhaoyuan Fang, Jie Li, Rares Ambrus, and Katerina Fragkiadaki.
\newblock Simple-{BEV}: What really matters for multi-sensor bev perception?
\newblock In \emph{IEEE International Conference on Robotics and Automation (ICRA)}, 2023.

\bibitem[Xie et~al.(2022)Xie, Yu, Zhou, Philion, Anandkumar, Fidler, Luo, and Alvarez]{Xie2022}
Enze Xie, Zhiding Yu, Daquan Zhou, Jonah Philion, Anima Anandkumar, Sanja Fidler, Ping Luo, and Jose~M Alvarez.
\newblock M\^{} 2bev: Multi-camera joint 3d detection and segmentation with unified birds-eye view representation.
\newblock \emph{arXiv preprint arXiv:2204.05088}, 2022.

\bibitem[Diaz-Zapata et~al.(2023)Diaz-Zapata, Sierra~Gonz{\'a}lez, Erkent, Dibangoye, and Laugier]{Diazzapata2023}
Manuel~Alejandro Diaz-Zapata, David Sierra~Gonz{\'a}lez, {\"O}zg{\"u}r Erkent, Jilles Dibangoye, and Christian Laugier.
\newblock {LAPTNet-FPN: Multi-scale LiDAR-aided Projective Transform Network for Real Time Semantic Grid Prediction}.
\newblock In \emph{{ICRA 2023 - IEEE International Conference on Robotics and Automation}}, pages 1--6, Londres, United Kingdom, May 2023. {IEEE Robotics and Automation Society}, {IEEE}.
\newblock URL \url{https://hal.science/hal-03980399}.

\bibitem[Saha et~al.(2022)Saha, Mendez, Russell, and Bowden]{Saha2022}
Avishkar Saha, Oscar Mendez, Chris Russell, and Richard Bowden.
\newblock Translating images into maps.
\newblock In \emph{2022 International Conference on Robotics and Automation, {ICRA} 2022, Philadelphia, PA, USA, May 23-27, 2022}, pages 9200--9206. {IEEE}, 2022.
\newblock \doi{10.1109/ICRA46639.2022.9811901}.

\bibitem[Fakoorian et~al.(2023)Fakoorian, Otsu, Khattak, Palieri, and Agha-mohammadi]{Rose23}
Seyed Fakoorian, Kyohei Otsu, Shehryar Khattak, Matteo Palieri, and Ali-akbar Agha-mohammadi.
\newblock Rose: Robust state estimation via online covariance adaption.
\newblock In Aude Billard, Tamim Asfour, and Oussama Khatib, editors, \emph{Robotics Research}, pages 452--467, Cham, 2023. Springer Nature Switzerland.
\newblock ISBN 978-3-031-25555-7.

\bibitem[Nubert et~al.(2022)Nubert, Khattak, and Hutter]{Nubert22}
Julian Nubert, Shehryar Khattak, and Marco Hutter.
\newblock Graph-based multi-sensor fusion for consistent localization of autonomous construction robots.
\newblock In \emph{2022 International Conference on Robotics and Automation (ICRA)}, pages 10048--10054, 2022.
\newblock \doi{10.1109/ICRA46639.2022.9812386}.

\bibitem[Strudel et~al.(2021)Strudel, Pinel, Laptev, and Schmid]{Strudel21}
Robin Strudel, Ricardo~Garcia Pinel, Ivan Laptev, and Cordelia Schmid.
\newblock Segmenter: Transformer for semantic segmentation.
\newblock In \emph{2021 {IEEE/CVF} International Conference on Computer Vision, {ICCV} 2021, Montreal, QC, Canada, October 10-17, 2021}, pages 7242--7252. {IEEE}, 2021.
\newblock \doi{10.1109/ICCV48922.2021.00717}.

\bibitem[Overbye and Saripalli(2022)]{Overbye2021}
Timothy Overbye and Srikanth Saripalli.
\newblock G-vom: A gpu accelerated voxel off-road mapping system.
\newblock In \emph{2022 IEEE Intelligent Vehicles Symposium (IV)}, pages 1480--1486, 2022.
\newblock \doi{10.1109/IV51971.2022.9827107}.

\bibitem[Williams et~al.(2017)Williams, Wagener, Goldfain, Drews, Rehg, Boots, and Theodorou]{Williams17}
Grady Williams, Nolan Wagener, Brian Goldfain, Paul Drews, James~M. Rehg, Byron Boots, and Evangelos~A. Theodorou.
\newblock Information theoretic mpc for model-based reinforcement learning.
\newblock In \emph{2017 IEEE International Conference on Robotics and Automation (ICRA)}, pages 1714--1721, 2017.
\newblock \doi{10.1109/ICRA.2017.7989202}.

\bibitem[Gibson et~al.(2023)Gibson, Vlahov, Fan, Spieler, Pastor, akbar Agha-mohammadi, and Theodorou]{Gibson2023multistep}
Jason Gibson, Bogdan Vlahov, David Fan, Patrick Spieler, Daniel Pastor, Ali akbar Agha-mohammadi, and Evangelos~A. Theodorou.
\newblock A multi-step dynamics modeling framework for autonomous driving in multiple environments, 2023.

\bibitem[Fakoorian et~al.(2021)Fakoorian, Santamaria{-}Navarro, Lopez, Simon, and Agha{-}mohammadi]{Fakoorian21}
Seyed~Abolfazl Fakoorian, Angel Santamaria{-}Navarro, Brett~Thomas Lopez, Dan Simon, and Ali{-}akbar Agha{-}mohammadi.
\newblock Towards robust state estimation by boosting the maximum correntropy criterion kalman filter with adaptive behaviors.
\newblock \emph{{IEEE} Robotics Autom. Lett.}, 6\penalty0 (3):\penalty0 5469--5476, 2021.
\newblock \doi{10.1109/LRA.2021.3073646}.

\bibitem[Contributors(2020)]{mmseg2020}
MMSegmentation Contributors.
\newblock {MMSegmentation}: Openmmlab semantic segmentation toolbox and benchmark.
\newblock \url{https://github.com/open-mmlab/mmsegmentation}, 2020.

\bibitem[{Duality Robotics, Inc.}(2023)]{Duality2023}
{Duality Robotics, Inc.}
\newblock Duality robotics - falcon pro simulator, 2023.
\newblock URL \url{https://www.duality.ai/product}.

\bibitem[Williams et~al.(2018)Williams, Drews, Goldfain, Rehg, and Theodorou]{Williams2018}
Grady Williams, Paul Drews, Brian Goldfain, James~M. Rehg, and Evangelos~A. Theodorou.
\newblock Information-theoretic model predictive control: Theory and applications to autonomous driving.
\newblock \emph{IEEE Transactions on Robotics}, 34\penalty0 (6):\penalty0 1603--1622, 2018.
\newblock \doi{10.1109/TRO.2018.2865891}.

\bibitem[Loshchilov and Hutter(2019)]{Ilya19}
Ilya Loshchilov and Frank Hutter.
\newblock Decoupled weight decay regularization.
\newblock In \emph{7th International Conference on Learning Representations, {ICLR} 2019, New Orleans, LA, USA, May 6-9, 2019}. OpenReview.net, 2019.
\newblock URL \url{https://openreview.net/forum?id=Bkg6RiCqY7}.

\bibitem[Smith and Topin(2017)]{Leslie17}
Leslie~N. Smith and Nicholay Topin.
\newblock Super-convergence: Very fast training of residual networks using large learning rates.
\newblock \emph{CoRR}, abs/1708.07120, 2017.
\newblock URL \url{http://arxiv.org/abs/1708.07120}.

\end{thebibliography}

\begin{IEEEbiography}[{\includegraphics[width=1in,height=1.25in,clip,keepaspectratio]{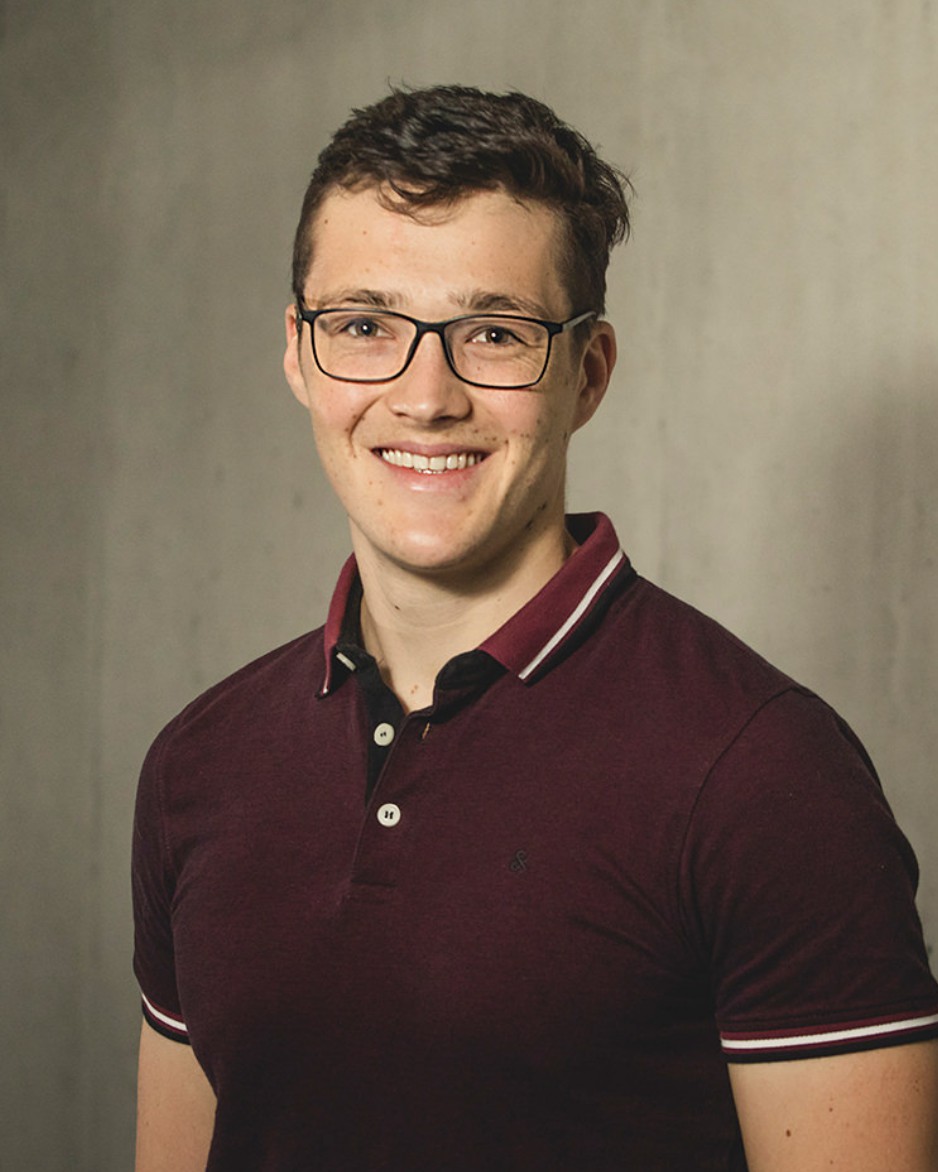}}]
{Jonas Frey}~(Student Member, IEEE)~is a PhD student in the Robotic Systems Lab at ETH Zurich. He received his M.Sc. in Robotics, Systems \& Control in 2021 from ETH Zurich. He is also affiliated with the Max Planck Institute through the MPI ETH Center for Learning Systems. His research interests lie in the field of perception, navigation and locomotion, and how it can be used for the deployment of mobile robotic systems.
\end{IEEEbiography} 
\begin{IEEEbiography}[{\includegraphics[width=1in,height=1.25in,clip,keepaspectratio]{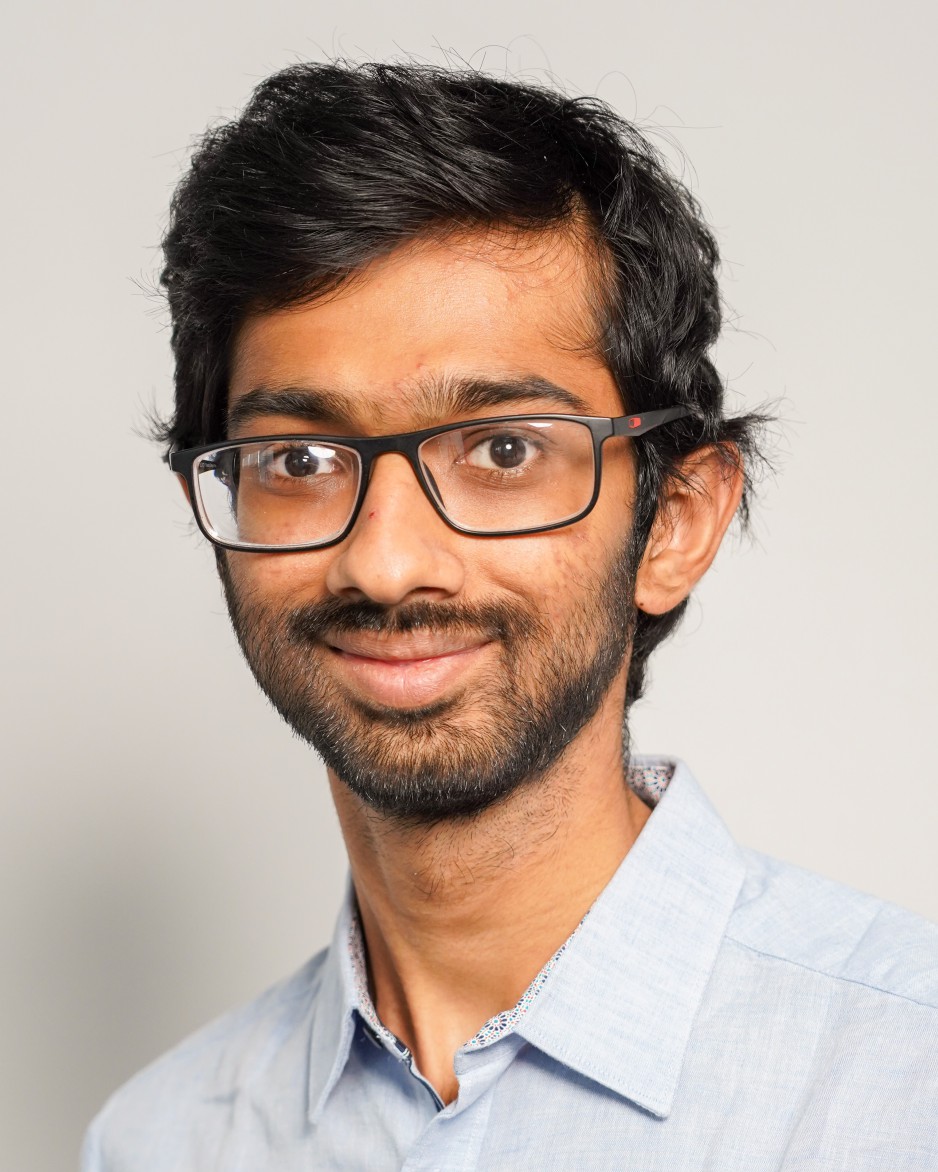}}]
{Manthan Patel}~(Student Member, IEEE)~is a research engineer at the Robotic Systems Lab at ETH Zurich. He received his M.Sc. in Robotics, Systems \& Control in 2024 from ETH Zurich, prior to which he received a BS in Mechanical Engineering from IIT Kharagpur. His research interests lie in various aspects of field robotics with a particular focus on perception for robots.
\end{IEEEbiography} 
\begin{IEEEbiography}[{\includegraphics[width=1in,height=1.25in,clip,keepaspectratio]{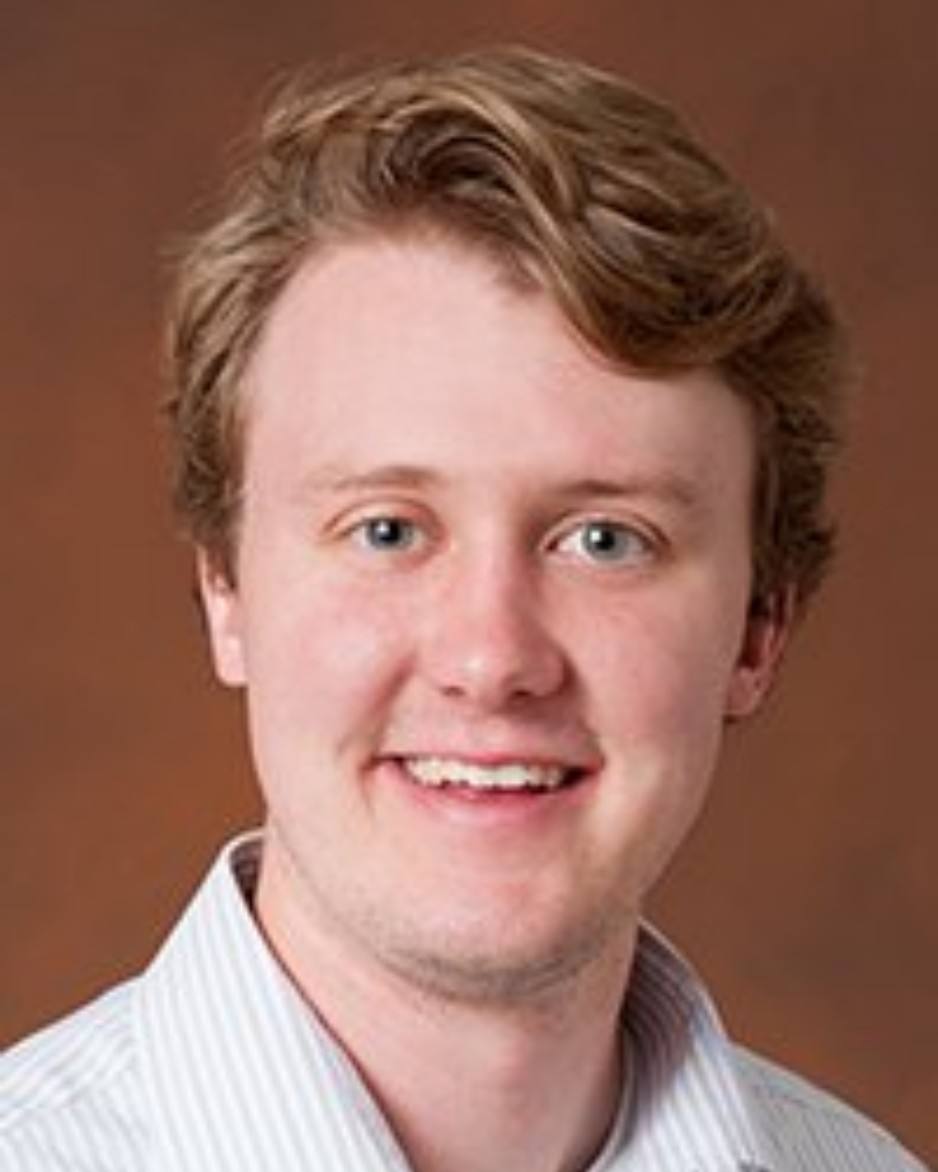}}]
{Deegan Atha}~is a Robotics Technologist within the Perception Systems Group of the Mobility and Robotic Systems Section at NASA's Jet Propulsion Laboratory. He received his B.S. degree from Purdue University in electrical engineering and his M.S. in computer science from the Georgia Institute of Technology. His research focuses on the infusion of robotic perception and learning into autonomous systems operating in unstructured environments. He is currently the Perception Lead for JPL's team in the DARPA RACER project. He has previously served as Principal Investigator for the ShadowNav task developing absolute localization methods for long-distance Lunar autonomy operating in darkness.

\end{IEEEbiography} 
\begin{IEEEbiography}
[{\includegraphics[width=1in,height=1.25in,clip,keepaspectratio]{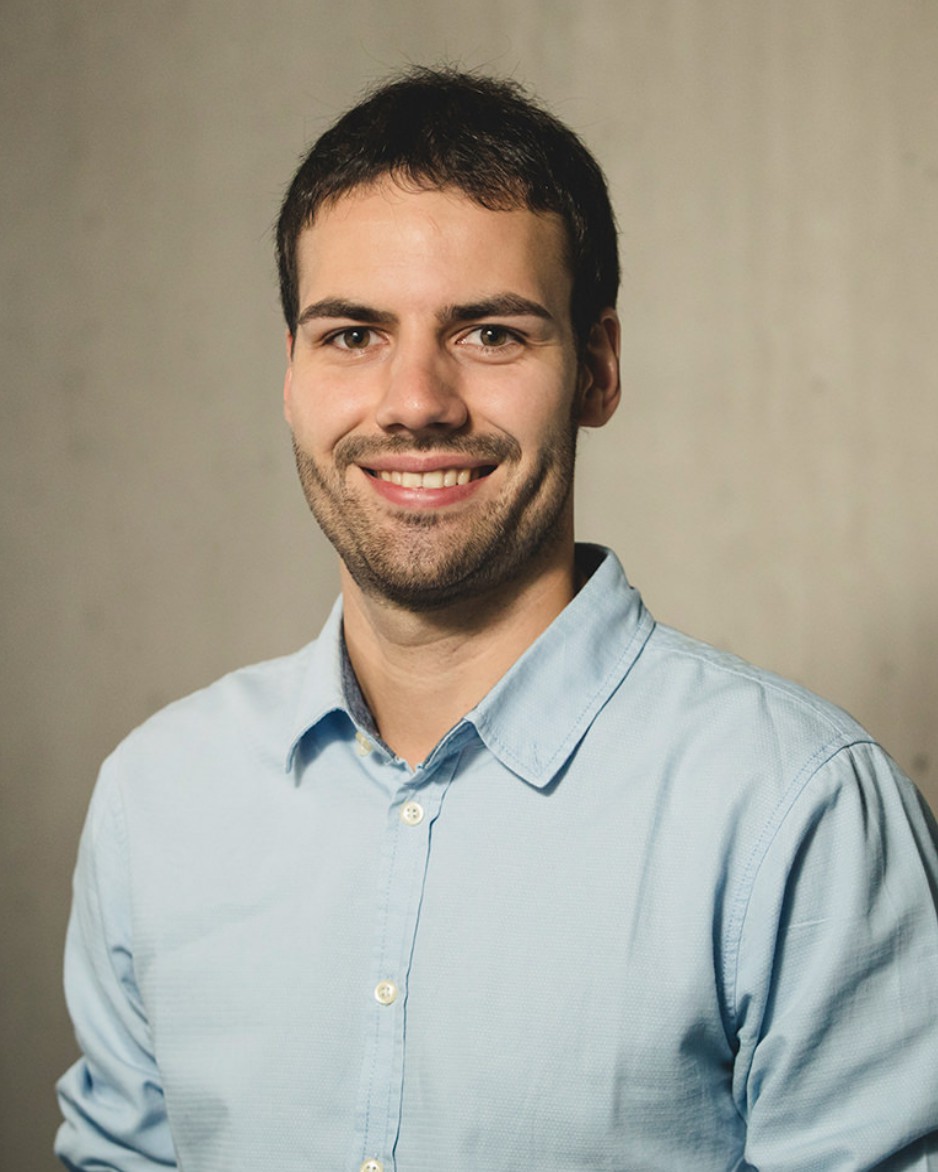}}]
{Julian Nubert}~(Student Member, IEEE)~is a PhD student in the Robotic Systems Lab at ETH Zurich. He received his M.Sc. in Robotics, Systems \& Control in 2020 from ETH Zurich. He is affiliated with the Max Planck Institute through the MPI ETH Center for Learning Systems. His research interests lie in robust robot perception and how it can be used to deploy mobile robotic systems. Julian received the ETH silver medal and was awarded the Willi-Studer-Price for his accomplishments during his master studies.
\end{IEEEbiography} 
\begin{IEEEbiography}[{\includegraphics[width=1in,height=1.25in,clip,keepaspectratio]{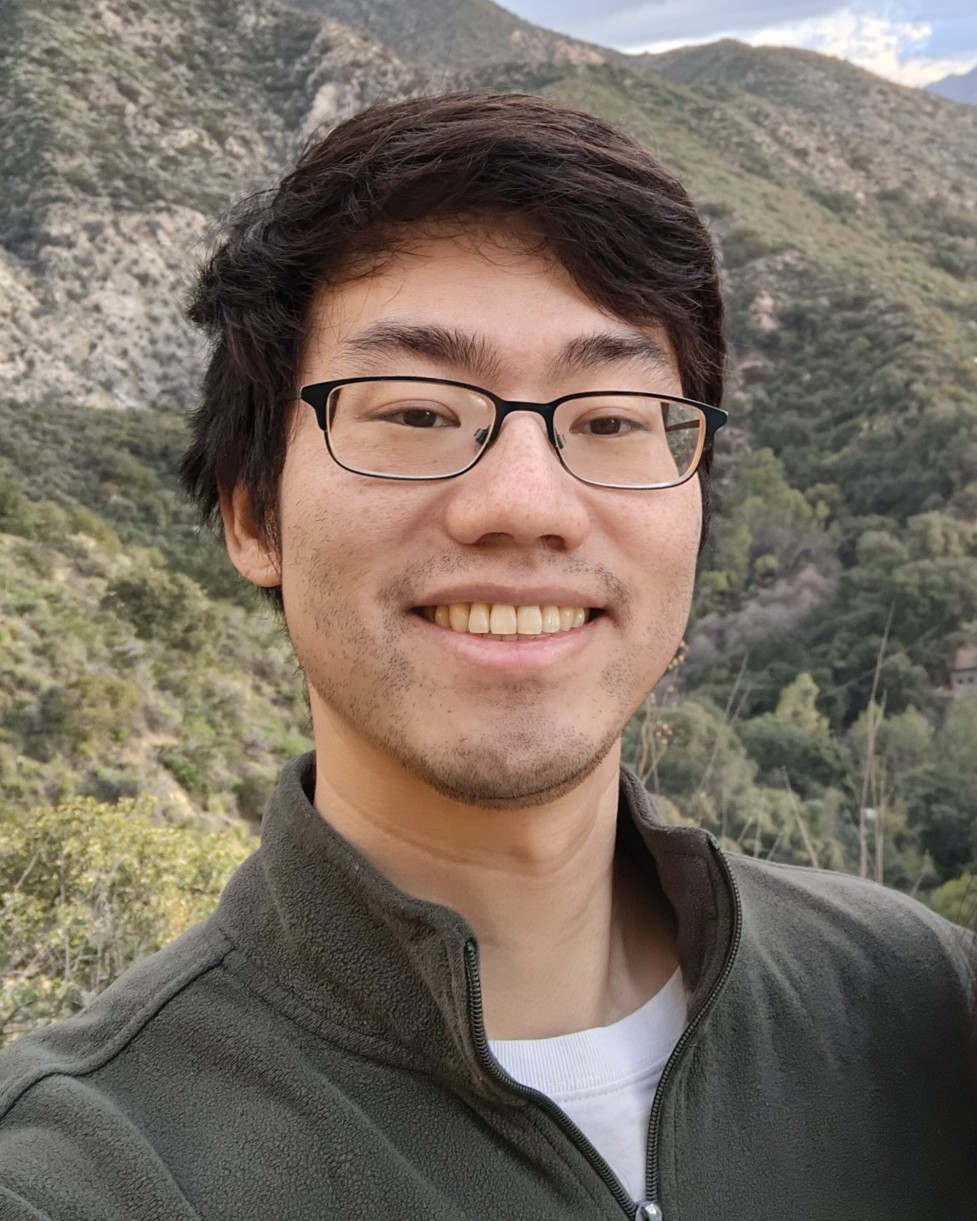}}]
{David Fan}~(Member, IEEE)~is a robotics researcher and technologist.  He received his Ph.D. from Georgia Institute of Technology in Robotics in 2021.  His research covers perception, planning, and controls using machine learning techniques, with a focus on safety-critical applications and fielding robots in challenging real-world scenarios.  He worked as a Research Associate at NASA Jet Propulsion Laboratory from 2019-2023, on the DARPA Subterranean Challenge and the DARPA RACER program.  He currently serves as Chief Technology Officer at Field AI, Inc.
\end{IEEEbiography} 

\afterpage{\clearpage}

\hspace{-0.44cm}
\noindent
\begin{minipage}[t]{\columnwidth-0.07cm}

\begin{IEEEbiography}[{\includegraphics[width=1in,height=1.25in,clip,keepaspectratio]{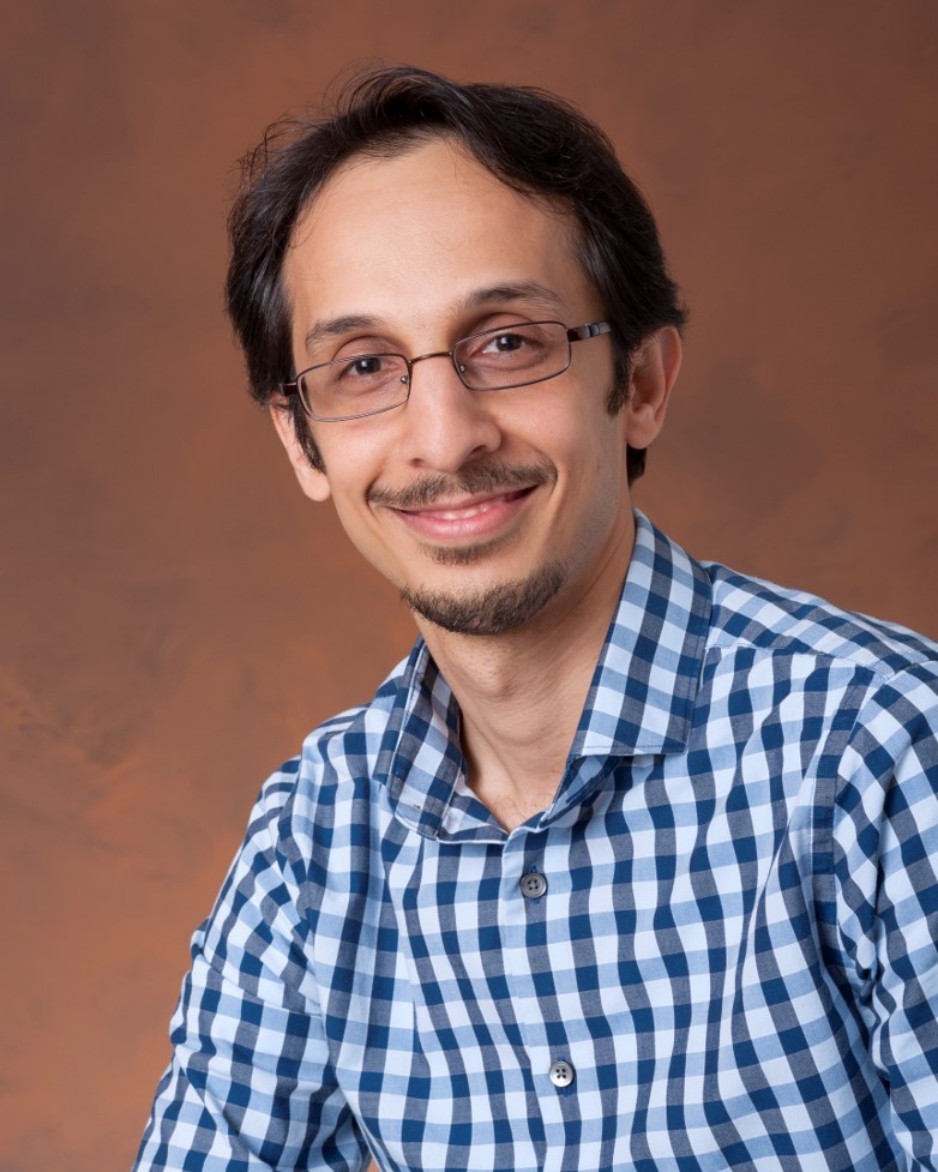}}]
{Ali Agha}~(Member, IEEE)~Ali Agha (Member, IEEE) is a co-founder of Field AI, a company dedicated to advancing next-generation robotic autonomy in complex and offroad environments. Before founding Field AI, Dr. Agha served as a Technologist and Group Leader at NASA-JPL and Caltech. During his tenure at JPL and Caltech, he was the principal investigator for and led NASA's team in several flagship autonomy-focused projects, including the DARPA Subterranean Challenge, DARPA Racer, and the prototype Mars Helicopter-Rover coordinated autonomy. Prior to his work at JPL, Dr. Agha was with MIT and Qualcomm Research, leading technical efforts in perception and planning for autonomous robotic vehicles.
\end{IEEEbiography} 

\begin{IEEEbiography}[{\includegraphics[width=1in,height=1.25in,clip,keepaspectratio]{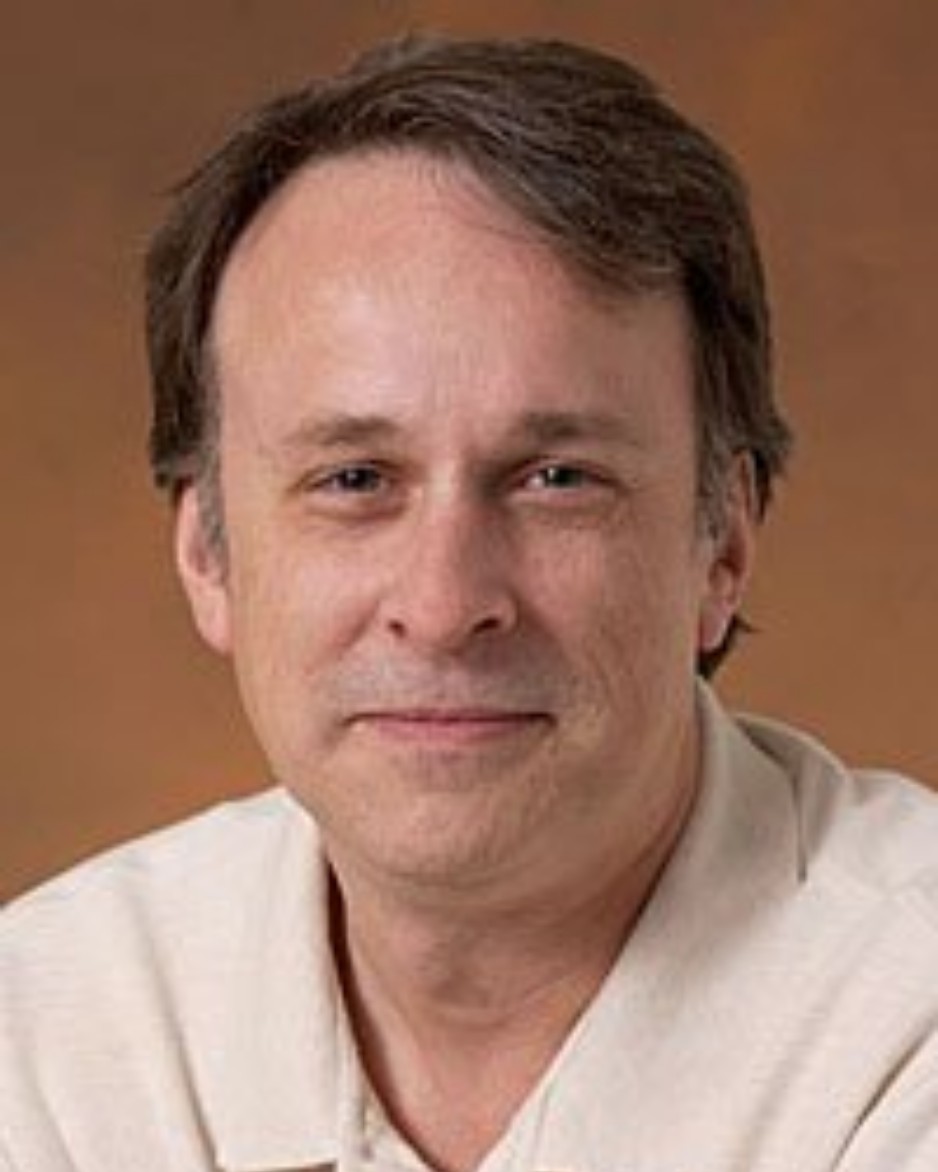}}]
{Curtis Padgett}~is the Supervisor for the Perception Systems Group and a Principal in the Robotics Section at NASA’s Jet Propulsion Laboratory where he has worked on machine vision problems for over 30 years. He leads research efforts focused on aerial and maritime imaging problems including: navigation support for landing and proximity operations; path planning for sea surface vehicles using COLREGs; automated, real-time recovery of structure from motion; precision geo-registration of imagery; automated landmark generation and mapping for surface relative navigation; stereo image sea surface sensing for navigation on water and image based, multi-platform contact range determination. He has a Ph.D. in Computer Science from the University of California at San Diego. His research interests include pattern recognition, image-based reconstruction, and mapping.
\end{IEEEbiography} 
\begin{IEEEbiography}[{\includegraphics[width=1in,height=1.25in,clip,keepaspectratio]{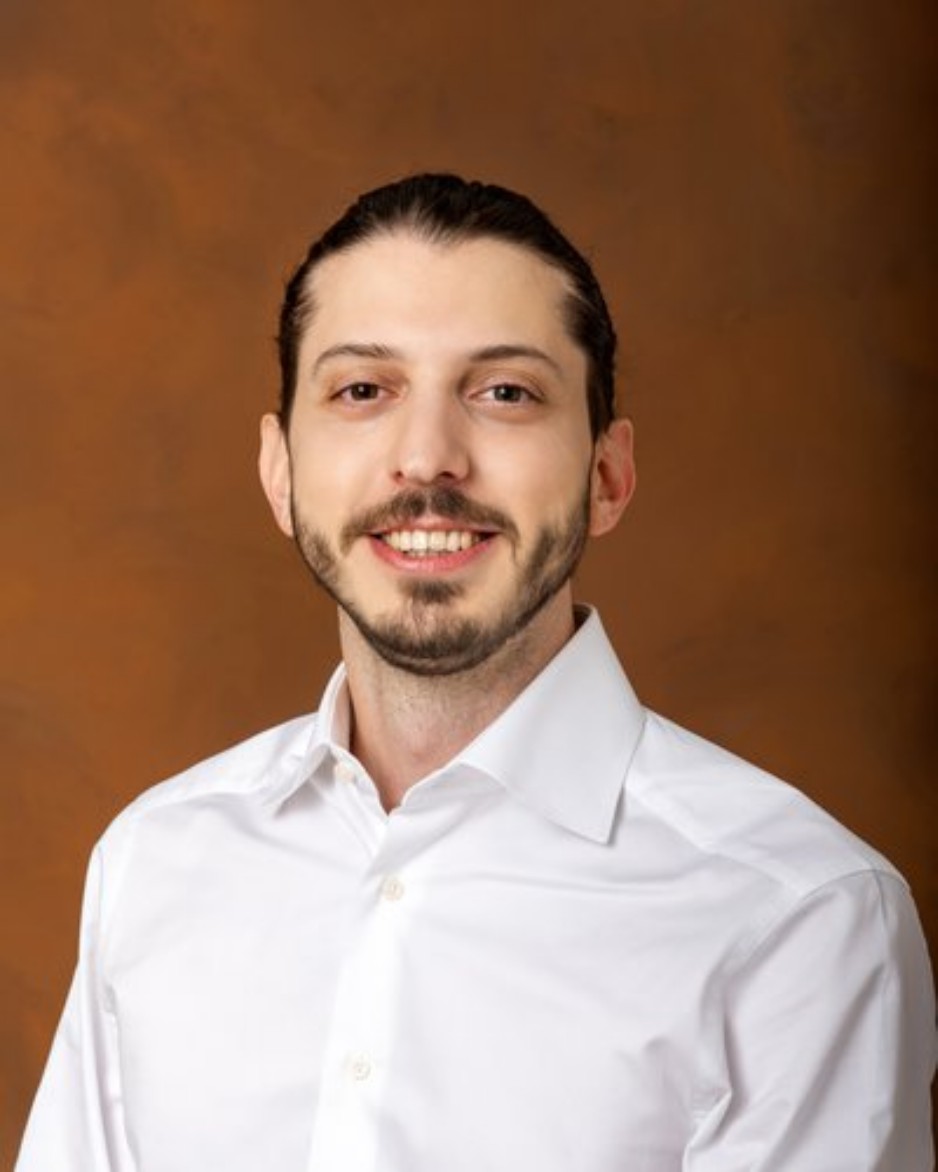}}]
{Patrick Spieler}~(Member, IEEE)~is a Robotics Technologist within the Aerial Mobility Group at NASA Jet Propulsion Laboratory. He received his M.S. (2017) in Robotics and Autonomous Systems and B.S. (2014) in Microengineering from the Swiss Federal Institute of Technology, Lausanne (EPFL). Currently, he is the principal investigator of the JPL team for the DARPA RACER project. Previously, he was a research engineer in the Autonomous Robotics and Control Lab (ARCL) at Caltech where he led the Autonomous Flying Ambulance project and Leonardo, the first flying-walking robot. Before that he worked at iRobot and Astrocast, a space company building communication satellites.
\end{IEEEbiography} 
\end{minipage}

\begin{minipage}[t]{\columnwidth-0.44cm}

\begin{IEEEbiography}[{\includegraphics[width=1in,height=1.25in,clip,keepaspectratio]{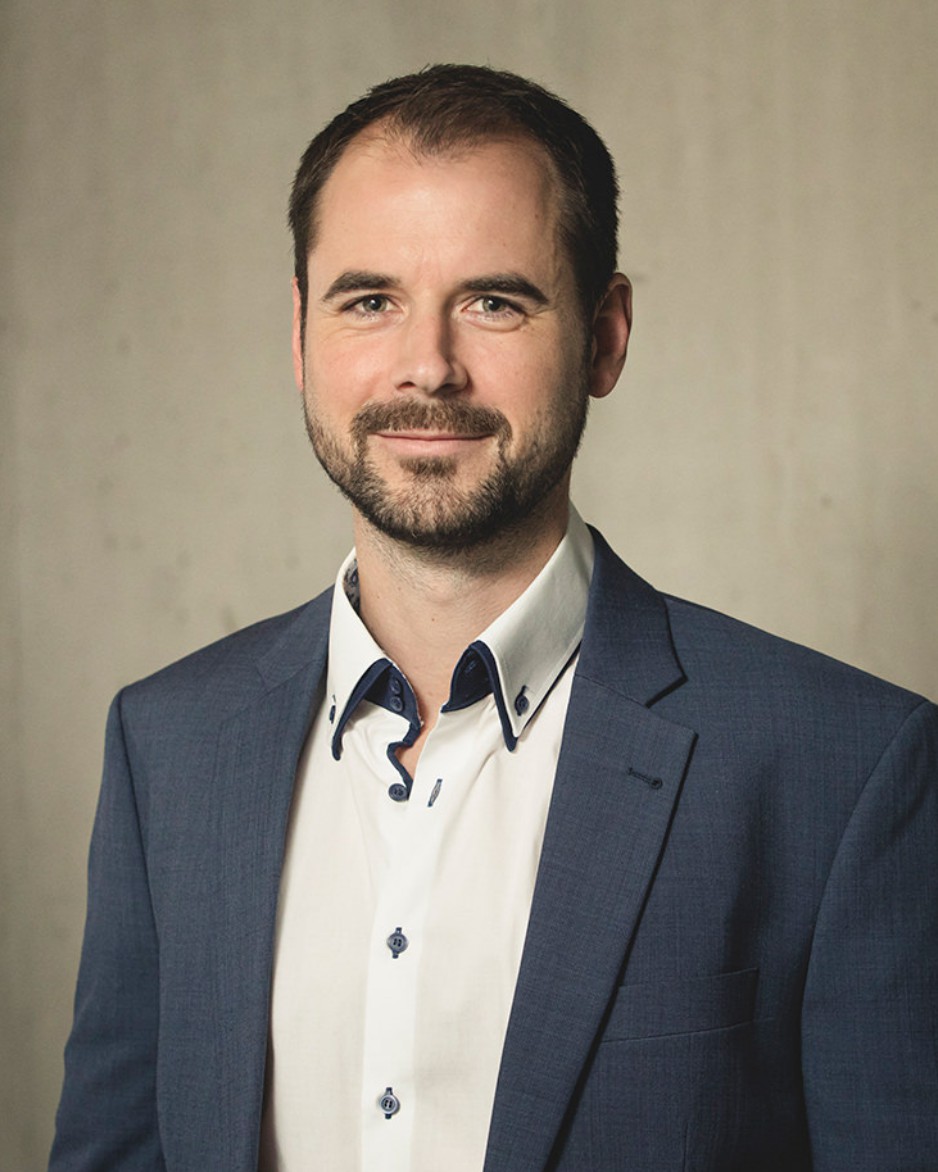}}]
{Marco Hutter}~(Member, IEEE)~is Associate Professor for Robotic Systems and Director of the Center for Robotics at ETH Zurich. He received his M.Sc. and PhD from ETH Zurich in 2009 and 2013 in the field of design, actuation, and control of legged robots. His research interests are in the development of novel machines and actuation concepts together with the underlying control, planning, and machine learning algorithms for locomotion and manipulation. Marco is the recipient of an ERC Starting Grant, PI of the NCCRs robotics, automation, and digital fabrication, winner of the DARPA SubT Challenge, and a co-founder of several ETH Startups such as ANYbotics and Gravis Robotics. He is also the Director of the Boston Dynamics AI Institute Zurich office.
\end{IEEEbiography} 

\begin{IEEEbiography}[{\includegraphics[width=1in,height=1.25in,clip,keepaspectratio]{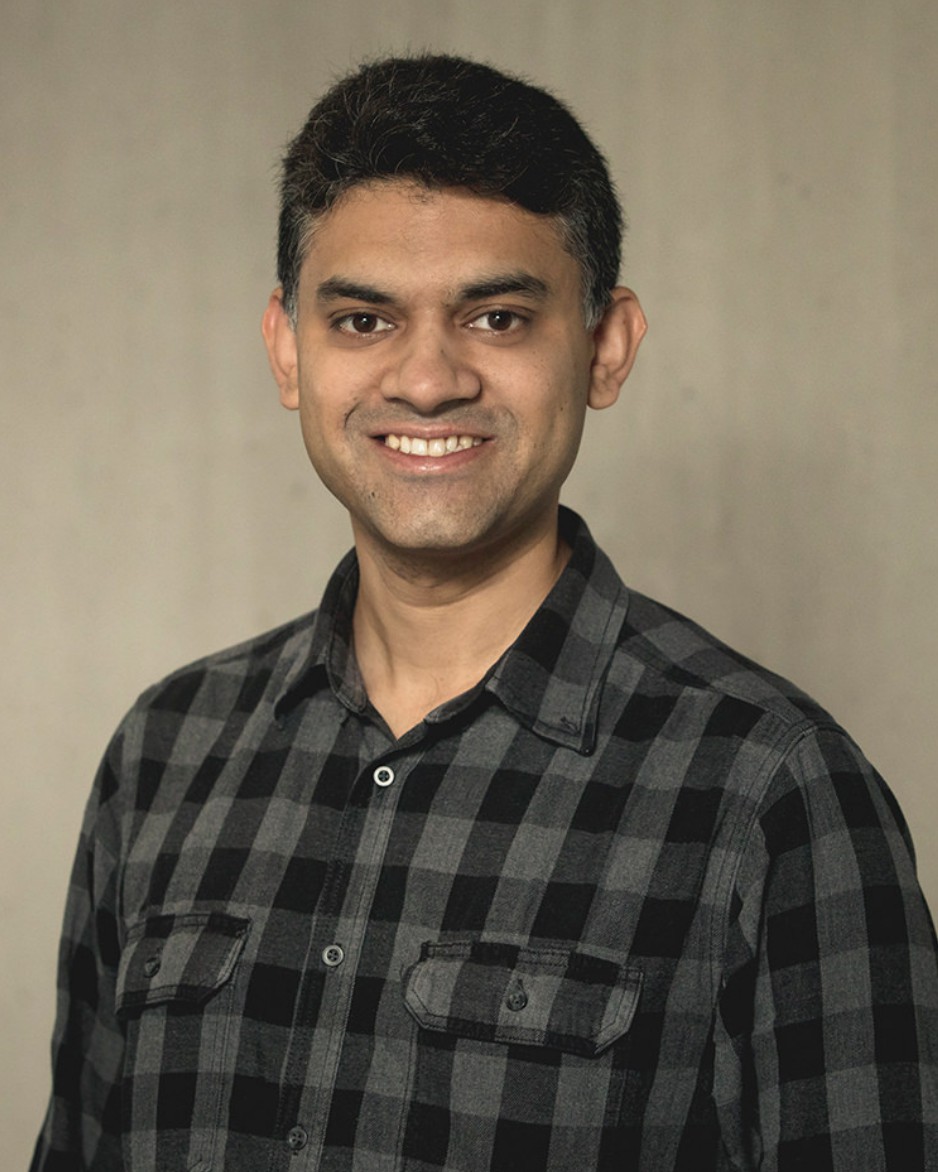}}]
{Shehryar Khattak}~(Member, IEEE)~Shehryar is a Robotics Technologist in the Perception Systems Group at NASA's Jet Propulsion Laboratory (JPL). His work focuses on enabling resilient robot autonomy in complex environments through multi-sensor information fusion. Currently, he is the Principal Investigator (PI) for the Multi-robot Autonomous Intelligent Search and Rescue task at JPL. He previously served as the perception lead for JPL's team in the DARPA RACER project and for Team CERBERUS, winners of the DARPA Subterranean Challenge. Before joining JPL, Shehryar was a postdoctoral researcher at ETH Zurich. He earned his Ph.D. (2019) and M.S. (2017) in Computer Science from the University of Nevada, Reno Additionally, he holds an M.S. in Aerospace Engineering from KAIST (2012) and a B.S. in Mechanical Engineering from GIKI (2009).
\end{IEEEbiography}
\end{minipage}

\end{document}